\documentclass[a4paper,fleqn]{cas-dc}

\usepackage[authoryear]{natbib}
\usepackage{amsmath,amssymb,amsfonts}
\usepackage{threeparttable}
\usepackage{booktabs}
\usepackage{multirow}

\definecolor{citeblue}{RGB}{0,114,188}
\hypersetup{
    citecolor=citeblue,
    linkcolor=citeblue,
}

\providecommand{\appendices}{\appendix}

\begin{document}

\let\WriteBookmarks\relax
\def\floatpagepagefraction{1}
\def\textpagefraction{.001}

\shortauthors{Y. Zhang et~al.}

\title [mode = title]{Seeing Globally, Refining Locally: Global Visual Guidance and Local Ultrasound Cues for Robust Freehand 3-D Ultrasound Reconstruction}
\tnotemark[1]
\tnotetext[1]{This work involved human subjects in its research. Approval of all ethical and experimental procedures and protocols was granted by Institutional Review Board, No. EA250796, Declaration of Helsinki.}

\author[1,2,3]{Yameng Zhang}
\ead{yamengzhang@mrc-cuhk.com}
\credit{Conceptualization, Methodology, Software, Data curation, Investigation, Formal analysis, Validation, Writing – original draft}

\author[2,3]{Zhongyu Chen}
\credit{Data curation, Validation, Writing – review and editing}

\author[1]{Dianye Huang}
\credit{Conceptualization, Methodology, Writing – review and editing}

\author[3]{Xiangyu Chu}
\credit{Resources, Writing – review and editing}

\author[2,3]{K. W. Samuel Au}
\credit{Resources, Funding acquisition, Writing – review and editing}

\author[1]{Zhongliang Jiang}
\cormark[1]
\ead{zljiang@hku.hk}
\credit{Supervision, Conceptualization, Methodology, Funding acquisition, Project administration, Writing – review and editing}

\affiliation[1]{organization={Department of Mechanical Engineering, The University of Hong Kong},
                country={Hong Kong SAR, China}}

\affiliation[2]{organization={Multi-Scale Medical Robotics Center},
                addressline={AIR@InnoHK},
                country={Hong Kong SAR, China}}

\affiliation[3]{organization={Department of Mechanical and Automation Engineering, The Chinese University of Hong Kong},
                country={Hong Kong SAR, China}}


\cortext[cor1]{Corresponding author}

\begin{abstract}
Freehand 3-D ultrasound (US) imaging has attracted increasing attention owing to its intuitive volumetric visualization, ease of use, and low cost. However, accurate 3-D reconstruction critically depends on stable probe pose estimation, yet existing trackerless methods remain susceptible to accumulated pose errors, particularly over long scanning trajectories. To address this limitation, we propose a global-to-local pose estimation framework that exploits external camera observations for globally stable localization and B-mode US images for anatomy-aware local refinement. Specifically, the framework comprises a dual-camera branch that performs contextual feature aggregation across camera views and temporal observations to estimate a globally consistent probe trajectory, and a B-mode branch that performs anatomical feature aggregation from sequential US images to capture tissue-dependent local motion cues. A cross-modal fusion module subsequently integrates the contextual camera features and anatomical US features to predict pose residuals and refine the camera-derived estimates in the transformation space. Furthermore, a multi-scale pose loss constrains relative motion over multiple temporal horizons to suppress accumulated drift during extended scans. The proposed framework is validated on phantom and \textit{in vivo} datasets. On two in-house datasets (FUSION-J and FUSION-L) collected using different machines, the proposed US~+~Dual-Cam model reduces average trajectory drift to $1.67$~mm and $1.29$~mm, representing improvement of $16.50\%$ and $27.12\%$, respectively, over a strong dual-camera baseline, while substantially outperforming US-only pose estimation ($>13$~mm drift). In \textit{in vivo} forearm arteries reconstruction, it achieves Hausdorff distances of $1.58$~mm, demonstrating the effectiveness of the proposed method on real clinical scenarios.
\end{abstract}



\let\printorcid\relax

\maketitle

\section{Introduction}
\label{sec:introduction}

Medical ultrasound (US) is widely used in clinical practice owing to its low cost, real-time imaging capability, and lack of ionizing radiation \citep{jiang2023robotic}. It has been broadly adopted in diagnostic imaging \citep{komatsu2025establishment}, vascular assessment \citep{yates2025improving}, and image-guided interventions \citep{huang2025vibnet}. Nevertheless, interpreting two-dimensional (2-D) US images remains challenging, because speckle artifacts, acoustic shadowing, and echo interference can obscure anatomical structures, making spatial interpretation highly dependent on operator experience. In contrast, three-dimensional (3-D) US imaging provides more intuitive volumetric information and is therefore valuable for diagnosis, anatomical visualization and assessment, and interventional guidance.

\begin{figure}[t]
\centering
\includegraphics[width=0.48\textwidth]{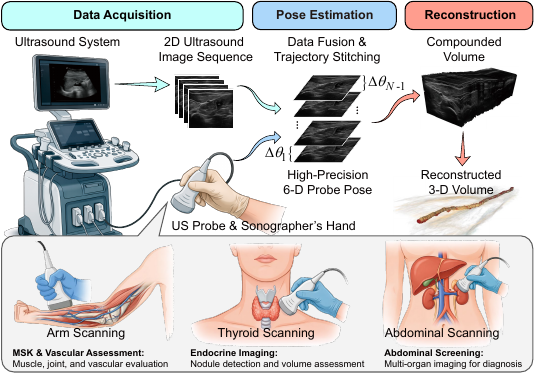}
\caption{Illustration of a freehand 3-D US imaging system using a 2-D US probe. The acquired 2-D images can be stacked to reconstruct 3-D volumes based on positional tracking data obtained from tracking devices or pose estimation algorithms.}%
\label{fig:teaser}
\end{figure}

Methods for generating 3-D US volumes can be broadly divided into two categories: matrix-array probe imaging, which directly acquires native 3-D volumes \citep{lee2023panoramic, bureau2023three}, and tracking-based 3-D imaging, which reconstructs volumetric data from tracked 2-D US sequences \citep{zhang2026navigation, adriaans2024trackerless}. Although matrix-array probes can acquire volumetric data directly, their high cost, complex signal interference, and limited field of view (FoV) restrict their widespread use in routine clinical and point-of-care scenarios \citep{wang2023multiplexed, hu2023wearable}. Tracking-based 3-D US imaging, as shown in Fig. \ref{fig:teaser}, offers a practical alternative by reconstructing a 3-D volume from sequential 2-D B-mode frames acquired with a standard transducer. In this approach, each 2-D image plane is placed in a common 3-D coordinate system according to its corresponding probe pose; hence, accurate probe pose estimation is essential for geometrically faithful reconstruction \citep{tehrani2022lateral}.

To obtain accurate probe pose information, conventional freehand systems commonly rely on precise external tracking devices, such as optical tracking systems \citep{verhoef2025freehand,zhang2026navigation} or electromagnetic (EM) tracking systems \citep{li2023automatic, wang2023virtual}. While these systems can provide accurate measurements under controlled conditions, they increase hardware cost, and their performance may be degraded by electromagnetic interference, optical occlusion, calibration errors, and workflow constraints. Robotic ultrasound systems can likewise provide accurate probe pose information and enable controlled scanning \citep{huang2024robot, jiang2021autonomous, jiang2023dopus}, but their cost and operational complexity limit their accessibility in many clinical environments.

These limitations have motivated a large body of work on sensorless, image-based, and camera-based freehand 3-D US reconstruction, which we review in Section~\ref{sec:related_work}. Despite substantial progress, existing approaches remain vulnerable to image degradation, visual ambiguity, and particularly, the accumulated drift over long freehand sweeps. Purely US image-based methods estimate probe motion from B-mode speckle patterns and inertial cues, but are susceptible to error accumulation, often resulting in average trajectory drift exceeding $5$~mm \citep{li2025tus}. Conversely, camera-based methods provide globally stable observations of the external scene but lack access to the internal anatomical information contained in ultrasound images. This complementary nature motivates the fusion of the two modalities, enabling globally consistent pose estimation while preserving anatomically accurate local motion \citep{men2023gaze}, particularly during long freehand sweeps in which accumulated errors can lead to pronounced trajectory drift.

In this paper, we propose a camera-guided residual pose refinement framework for freehand 3-D US reconstruction. During freehand scanning, the system synchronously acquires dual-camera RGB image sequences and B-mode US image sequences. The camera branch first provides a coarse but globally stable probe pose estimate from dual-camera observations, while the ultrasound branch extracts motion-aware latent representations from sequential B-mode frames to characterize tissue appearance changes and probe-tissue interaction. These two sources of information are then integrated by a cross-modal fusion module, which predicts a residual pose correction to compensate for local errors in the camera-derived estimate. To improve both frame-level accuracy and long-horizon trajectory consistency, we further introduce a multi-scale pose loss that supervises motion errors over multiple temporal horizons. In this way, the proposed framework combines the global stability of external visual tracking with the local sensitivity of ultrasound motion cues to achieve more accurate and robust freehand 3-D US reconstruction. The main contributions of this work are summarized as follows:

\begin{itemize}
\item We propose a global-to-local pose estimation method for robust freehand 3-D US reconstruction that, for the first time, integrates external camera-based visual cues with internal US image-domain motion cues to refine camera-derived probe poses. The proposed framework improves pose estimation accuracy and produces higher-quality 3-D US volumes, particularly under challenging practical conditions, including image degradation, visual ambiguity, and long or back-and-forth freehand sweeps.

\item We design an ultrasound-based pose estimation branch that extracts spatiotemporal representations from sequential B-mode frames. A multi-scale pose loss supervises relative motion over varying temporal horizons, with rotational error measured by the geodesic distance on $SO(3)$ to capture the intrinsic shortest-angle discrepancy. Besides, translation and rotation terms are balanced using learnable uncertainty weights. Combined with the correlation and triplet losses, the proposed US image-based pose estimation branch alone achieves state-of-the-art performance on both the FUSION-J and FUSION-L\footnote{The URL will be added after publication.} datasets and the public TUS-REC2025 \footnote{\url{https://zenodo.org/records/15699959}} dataset.

\end{itemize}

Finally, we validate the proposed method on TUS-REC2025 for benchmarking and on two in-house datasets (FUSION-J and FUSION-L) collected using different machines, demonstrating its potential in real scenarios. On FUSION-J and FUSION-L, the proposed US~+~Dual-Cam model reduces the average drift to $1.67$~mm and $1.29$~mm, respectively. Experiments on phantom and \textit{in vivo} human forearm artery scans further yield reconstruction Hausdorff distances of $1.58$-$2.35$~mm and Dice scores of $0.69$-$0.87$, demonstrating improved probe pose estimation and high-quality 3-D reconstruction of both anatomical and phantom structures. We will release both the datasets and the source code upon acceptance.

The remainder of this paper is organized as follows. Section~\ref{sec:related_work} reviews related work on freehand 3-D US reconstruction. Section~\ref{sec:methods} formulates the problem and presents the proposed framework, including the camera-based pose estimation branch, the US-based pose estimation branch, the cross-modal residual refinement module, and the loss functions. Section~\ref{sec:experiments} describes the datasets, implementation details, evaluation metrics, and quantitative and qualitative reconstruction results. Sections~\ref{sec:discussion} and~\ref{sec:conclusion} discuss the main findings and conclude the paper, respectively.

\section{Related Work}
\label{sec:related_work}

\subsection{Conventional US Image-based Pose Estimation}
\label{subsec:sensorless}
To reduce the dependence on external tracking devices, sensorless and image-based freehand 3-D US reconstruction has been extensively investigated. Early methods estimated scan-plane motion by analyzing speckle decorrelation or by registering adjacent B-mode frames \citep{trahey1986speckle,tuthill1998automated,chang20033,gee2006sensorless,laporte2011learning}. These works demonstrated that B-mode images contain motion-related information, conveyed particularly through tissue speckle and local anatomical appearance. However, they relied heavily on hand-crafted speckle or image-similarity cues, which limited their robustness under out-of-plane motion, image noise, and long-sequence drift.

\subsection{Learning-based Ultrasound Pose Estimation}
\label{subsec:learning}

With the advent of deep learning, neural networks have been introduced to learn scan motion directly from US image sequences, with several studies additionally incorporating cost-effective inertial measurement unit (IMU) sensors to provide complementary motion cues. \citet{prevost20183d} pioneered this paradigm with an end-to-end convolutional neural network (CNN)-based framework that directly regresses inter-frame six-degree-of-freedom (6-DoF) probe motion from B-mode images, with optical-flow and IMU-assisted cues improving motion estimation robustness.

Following this formulation, subsequent studies have mainly sought to improve temporal modeling, feature representation, and loss design to reduce cumulative drift. \citet{guo2020sensorless} proposed DCL-Net to exploit temporal contextual information from multiple consecutive B-mode frames rather than estimating motion only from adjacent frame pairs. Their subsequent DC\textsuperscript{2}-Net further incorporated self-attention, a case-wise correlation loss, and a contrastive margin ranking loss to improve trajectory estimation and volume reconstruction \citep{guo2022ultrasound}. \citet{ning2022spatial} proposed USFormer, a hybrid Transformer-based method that integrates CNN-extracted US image features, Transformer-based sequence modeling, and IMU feature embedding to reduce cumulative drift in long-sequence reconstruction.

Several more recent methods further extend this direction by strengthening sequence modeling or introducing additional motion constraints. \citet{luo2022deep} proposed the MoNet series, which fuses B-mode image sequences with lightweight IMU-derived motion cues and progressively improves external-tracker-free reconstruction through temporal multi-branch fusion, online consistency constraints, and multimodal self-supervision \citep{luo2025monetv2}. \citet{li2023long} investigated long-term dependency in trackerless freehand 3-D US reconstruction and showed that incorporating distant frames through sequence encoding and multi-transformation prediction can substantially reduce accumulated reconstruction error. Their team also promoted community benchmarking by organizing the TUS-REC2024/2025 challenges and releasing public datasets and evaluation protocols \citep{li2025tus}.

Beyond temporal context, other studies have explored more discriminative motion representations. \citet{yan2024fine} proposed FiMA, which employs a ReMamba-based state space model to capture fine-grained spatio-temporal dependencies in US image sequences. By combining multi-directional sequence modeling, adaptive multi-IMU fusion, and online multimodal alignment, FiMA achieved stable reconstruction performance on arm and carotid datasets. \citet{dou2024sensorless} proposed PLPPI, a physics-guided sensorless method that incorporates a speckle-decorrelation-inspired correlation operator to model inter-frame motion without 3-D convolutions, improving both accuracy and computational efficiency. \citet{lee2025enhancing} proposed MoGLo-Net, which combines patch-wise correlation, global-local attention, and motion-aware supervision for freehand 3-D US motion estimation, and further extended freehand 3-D reconstruction to Power Doppler and photoacoustic vascular imaging. \citet{wilson2025dualtrack} proposed DualTrack, a dual-encoder framework that explicitly separates local speckle-related motion cues from global anatomical context, achieving state-of-the-art performance on the public TUS-REC benchmark.

Despite their methodological advances, these learning-based methods remain constrained by the nature of their inputs. The out-of-plane cues in B-mode images are implicit and often insufficient to fully constrain the 6-DoF probe pose, while speckle noise, anatomical deformation, probe-tissue interaction, and limited inter-frame overlap further complicate motion estimation. Consequently, accumulated drift over long freehand sweeps remains difficult to avoid, with average drift typically exceeding $5~\mathrm{mm}$ \citep{li2025tus}. Robust generalization across scanners, anatomies, and scanning protocols also remains an open problem. These limitations have motivated the community to explore more stable and explicit observations beyond noise-sensitive B-mode and IMU cues.

Along this direction, \citet{liang2026sensorless} embedded structured marker patterns into the coupling device, enabling pose estimation from explicit B-mode image patterns rather than from tissue speckle alone. Although this design reduces average drift to below $2$~mm, its dependence on a customized coupling pad and associated calibration may limit its scalability to longer or more flexible scan trajectories, motivating the use of external visual observations for probe motion estimation.

\subsection{External Vision-based Pose Estimation}
\label{subsec:camera}

External visual observations have also been explored for probe pose estimation. Early efforts explored probe-mounted cameras: \citet{sun2014probe} estimated 6-DoF probe motion by tracking natural skin features, achieving accumulated errors of $0.91 \pm 0.49$~mm in translation and $0.55^{\circ} \pm 0.17^{\circ}$ in rotation per $10$~mm of travel. This method remained susceptible to accumulated drift, translation-rotation ambiguity, and physiological motion.

More recent methods leverage stronger visual models and structured environments to improve accuracy. \citet{zhang2026mlrecon} proposed MLRecon, a markerless framework that tracks the 6-DoF probe pose using a single RGB-D camera together with vision foundation models, including FoundationPose \citep{wen2024foundationpose} and SAM~2-based segmentation \citep{ravi2025sam}, reporting average positional and rotational errors of $1.39$~mm and $0.71^\circ$; however, its dependence on continuous RGB-D visibility makes it sensitive to occlusion and viewpoint constraints. \citet{huang2025improving} instead estimated the global 6-DoF probe pose from two probe-mounted monocular cameras in a semi-structured marker environment, training a cross encoder-decoder in simulation and adapting it to real data with PoseNet to reduce accumulated drift to $2.03$~mm, although Sim2Real adaptation and marker-detection errors still limit robustness in complex clinical settings.

\citet{zhang2025freehand} formulated camera-based probe localization as a simulation-in-the-loop visual servoing problem, in which the simulated camera view is iteratively aligned with real observations to estimate the global 6-DoF probe pose. This design mitigates Sim2Real discrepancy and partial visual occlusion, achieving an average positional error below $2$~mm. Nevertheless, because camera-based methods observe the external scene rather than the scanned tissue, the anatomical and speckle-based motion cues embedded in US image sequences remain unexploited. This limitation directly motivates the dual-modal fusion framework proposed in this work.

\section{Methods}
\label{sec:methods}

\subsection{Problem Definition}
\label{subsec:problem_definition}

Consider a freehand scan consisting of $N$ synchronized observations. At frame $i$, the system acquires a B-mode ultrasound image $\mathbf{U}_i \in \mathbb{R}^{H_u \times L_u}$ and dual-camera RGB images $\mathbf{W}_i \in \mathbb{R}^{2 \times H_w \times L_w}$, where the first dimension of $\mathbf{W}_i$ indexes the two camera views. The spatial location of the corresponding ultrasound image plane is represented by a rigid transformation
\begin{equation}
\mathbf{T}_i =
\begin{bmatrix}
\mathbf{R}_i & \mathbf{t}_i \\
\mathbf{0}^{\top} & 1
\end{bmatrix}
\in SE(3),
\end{equation}
where $\mathbf{R}_i \in SO(3)$ and $\mathbf{t}_i \in \mathbb{R}^{3}$ denote the probe orientation and translation, respectively.

Given the synchronized ultrasound and camera sequences,
\begin{equation}
\mathcal{X}
=
\left\{
(\mathbf{U}_i,\mathbf{W}_i)
\right\}_{i=0}^{N-1},
\end{equation}
the objective is to estimate the full probe trajectory
\begin{equation}
\hat{\mathcal{T}}
=
\left\{
\hat{\mathbf{T}}_i
\right\}_{i=0}^{N-1}
=
f(\mathcal{X}),
\end{equation}
such that the estimated ultrasound image planes remain geometrically consistent with the ground-truth trajectory over the entire freehand sweep.

Although individual network components may predict frame-to-frame motion or operate on short temporal windows, these intermediate estimates are ultimately composed or fused to recover the global probe trajectory. Therefore, the primary objective is to reduce accumulated translational and rotational errors over the full scan:
\begin{equation}
\hat{\mathcal{T}}^{*}
=
\arg\min_{\hat{\mathcal{T}}}
\mathcal{E}_{\mathrm{traj}}
\left(
\hat{\mathcal{T}},
\mathcal{T}
\right),
\end{equation}
where $\mathcal{T}=\{\mathbf{T}_i\}_{i=0}^{N-1}$ denotes the ground-truth trajectory and $\mathcal{E}_{\mathrm{traj}}$ measures trajectory-level geometric discrepancy. An accurate trajectory enables the acquired 2-D ultrasound frames to be placed consistently in a common 3-D coordinate system, thereby supporting geometrically faithful volumetric reconstruction.

\subsection{Overall Framework}
\label{subsec:dual_stream_framework}

\begin{figure}[t]
\centering
\includegraphics[width=0.45\textwidth]{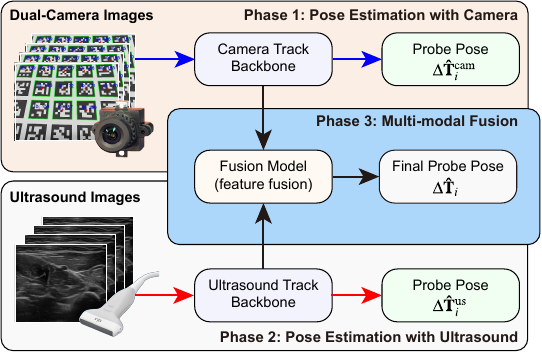}
\caption{Overview of the proposed framework. The camera and ultrasound backbones extract complementary motion cues from dual-camera RGB observations and B-mode image sequences, respectively, and their representations are fused to estimate the final probe pose.}
\label{fig:pipeline}
\end{figure}

To recover the full probe trajectory defined above, we adopt the global-to-local framework illustrated in Fig.~\ref{fig:pipeline}. The camera backbone extracts contextual features from dual-camera RGB observations, while the ultrasound backbone captures anatomical and motion-related features from sequential B-mode images. Their latent representations are subsequently fused to predict residual corrections to the camera-derived poses, yielding a refined trajectory over the entire freehand sweep.

\subsubsection{Contextual Feature Aggregation from Camera Images}
\label{subsec:camera_pipeline}

\begin{figure*}[htp!]
\centering
\includegraphics[width=0.8\textwidth]{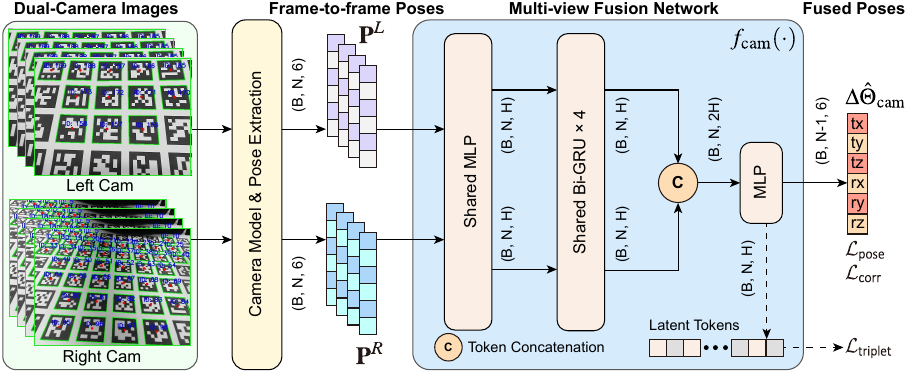}
\caption{Architecture of the camera contextual feature aggregation branch. AprilTag-based PnP estimates a coarse 6-DoF pose from each RGB view. A shared MLP and Bi-GRU aggregate temporal context within each relative motion sequence, after which the two view-specific representations are fused to predict camera-derived probe motion.}
\label{fig:cam_pose}
\end{figure*}

The camera branch aggregates temporal and cross-view context from synchronized dual-camera RGB sequences to obtain globally anchored visual pose information. As shown in Fig.~\ref{fig:cam_pose}, AprilTag-based pose estimates are first computed independently from the two camera views. The resulting relative motion sequences are then encoded using shared recurrent networks, and their view-specific features are fused to estimate camera-derived probe motion.

For each view, detected AprilTag corners establish 2-D-3-D correspondences with their known locations on the planar tag grid. Let $g$ and $c$ denote the tag-grid and camera coordinate systems, respectively. After image-point undistortion, the grid-to-camera transformation $\mathbf{T}_{g \rightarrow c}$ is estimated using the calibrated intrinsic matrix $\mathbf{K}$ and the PnP projection model
\begin{equation}
\tilde{\mathbf{u}}_j
\sim
\mathbf{K}
\left[
\mathbf{R}_{g \rightarrow c}
\mid
\mathbf{t}_{g \rightarrow c}
\right]
\tilde{\mathbf{X}}_j,
\label{eq:cam_pnp}
\end{equation}
where
$\tilde{\mathbf{X}}_j=[X_j,Y_j,Z_j,1]^{\top}$
and
$\tilde{\mathbf{u}}_j=[u_j,v_j,1]^{\top}$
denote the homogeneous coordinates of the $j$-th tag corner and its corresponding image observation, respectively.

The transformation $\mathbf{T}_{g \rightarrow c}$ is inverted and combined with the calibrated camera-to-ultrasound extrinsic transformation to express the ultrasound image-center pose in the tag-grid reference frame. The resulting absolute pose sequences provide coarse, globally referenced probe trajectories for subsequent residual refinement. Adjacent poses are also converted into left- and right-view relative motion sequences for contextual feature learning.

A zero vector is prepended to each relative motion sequence for temporal alignment, yielding
$\mathbf{P}^{L},\mathbf{P}^{R}\in\mathbb{R}^{B\times N\times6}$,
where $B$ denotes the batch size. A shared MLP is applied independently to the two sequences to embed them into a common feature space. A shared bidirectional gated recurrent unit (Bi-GRU) then aggregates bidirectional temporal context within each view. The resulting frame-wise features are concatenated and fused by an MLP head to integrate complementary evidence from the two viewpoints.

After discarding the output associated with the prepended zero vector, the branch produces
\begin{equation}
\Delta\hat{\boldsymbol{\Theta}}_{\mathrm{cam}}
=
f_{\mathrm{cam}}
\left(
\mathbf{P}^{L},
\mathbf{P}^{R}
\right)
\in
\mathbb{R}^{B\times(N-1)\times6},
\label{eq:cam_network}
\end{equation}
where $\Delta\hat{\boldsymbol{\Theta}}_{\mathrm{cam}}$ denotes the camera-derived relative pose sequence. The fused hidden features encode within-view temporal context and cross-view complementary information, and are retained as the contextual camera representation for subsequent dual-modal fusion. The corresponding absolute camera-derived poses are retained as the coarse global trajectory to be refined by the fusion module.

\subsubsection{Anatomical Feature Aggregation from US Images}
\label{subsec:us_pipeline}

\begin{figure*}[htp!]
\centering
\includegraphics[width=0.95\textwidth]{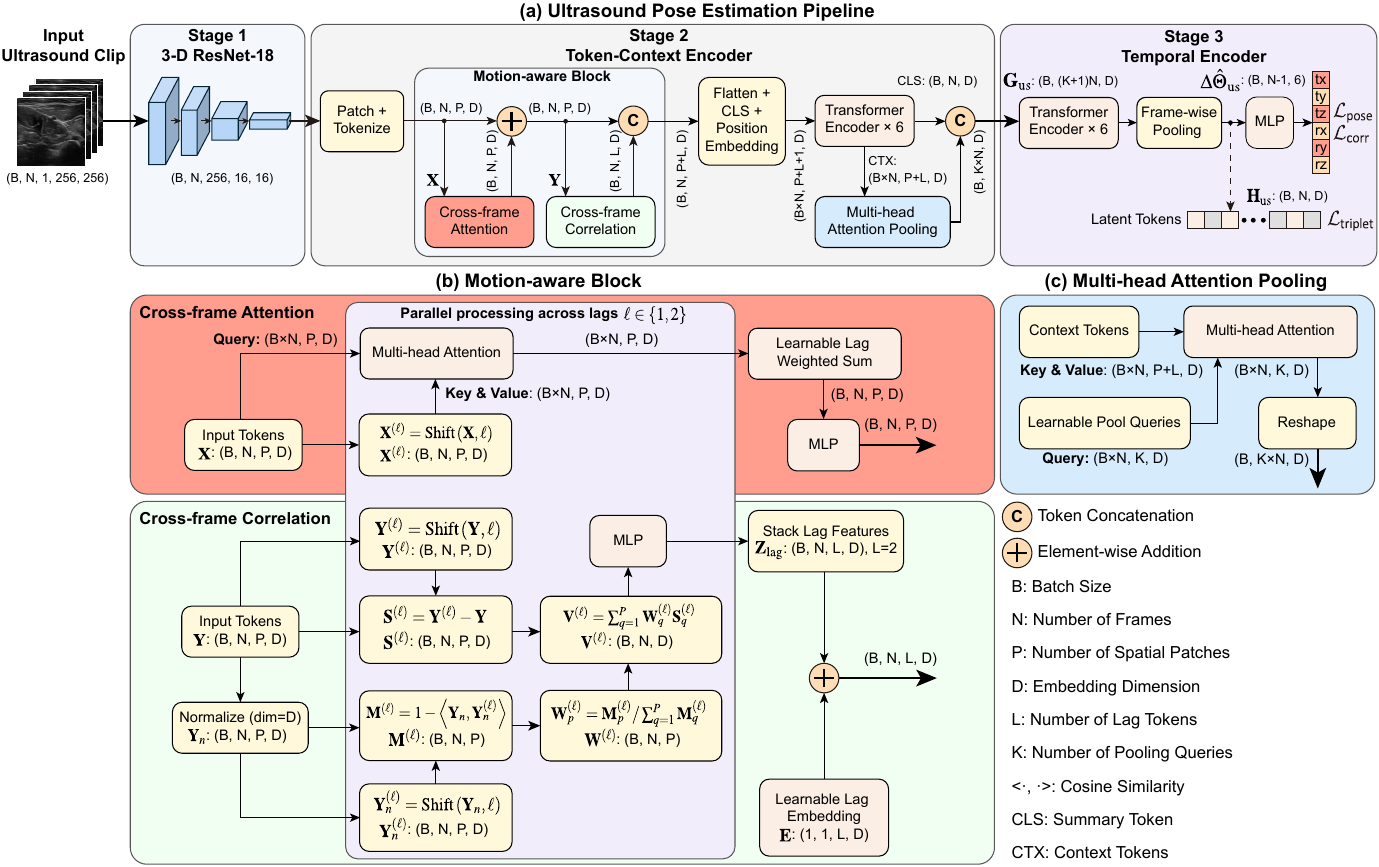}
\caption{Anatomical feature aggregation from sequential B-mode images. A 3-D ResNet-18 encodes local anatomical features, a motion-aware token-context encoder aggregates cross-frame appearance changes, and a temporal encoder integrates clip-level context for relative probe motion estimation.}
\label{fig:us_pose}
\end{figure*}

The ultrasound branch aggregates anatomical appearance and motion-related cues from sequential B-mode images to estimate relative probe motion and provide latent representations for subsequent cross-modal fusion. As shown in Fig.~\ref{fig:us_pose}, it consists of a 3-D ResNet-18 backbone, a token-context encoder, and a temporal encoder. Given an ultrasound clip
$\mathbf{U}\in\mathbb{R}^{B\times N\times 1\times H\times W}$,
the branch predicts
\begin{equation}
\Delta \hat{\boldsymbol{\Theta}}_{\mathrm{us}}
=
f_{\mathrm{us}}(\mathbf{U})
\in\mathbb{R}^{B\times (N-1)\times 6},
\end{equation}
where $B$ and $N$ denote the batch size and number of frames, respectively.

\paragraph{Stage 1: 3-D ResNet-18.}
A 3-D ResNet-18 backbone transforms the input clip into frame-aligned feature maps of size $(B,N,256,16,16)$. These features encode local anatomical appearance, speckle patterns, and short-range spatiotemporal context at a reduced spatial resolution.

\paragraph{Stage 2: token-context encoder.}
The feature maps are converted into frame-wise patch tokens
$\mathbf{X}\in\mathbb{R}^{B\times N\times P\times D}$,
where $P$ and $D$ denote the number of patches and embedding dimension, respectively. A motion-aware block then aggregates short-range anatomical variations through cross-frame attention and correlation.

For each temporal lag $\ell\in\{1,2\}$, the token sequence is shifted to obtain
$\mathbf{X}^{(\ell)}=\operatorname{Shift}(\mathbf{X},\ell)$.
Using $\mathbf{X}$ as queries and $\mathbf{X}^{(\ell)}$ as keys and values, cross-frame attention is computed:
\begin{equation}
\mathbf{A}^{(\ell)}
=
\operatorname{MHA}
\left(
\mathbf{X},
\mathbf{X}^{(\ell)},
\mathbf{X}^{(\ell)}
\right),
\end{equation}
and the lag-specific responses are combined with learnable weights:
\begin{equation}
\mathbf{Y}
=
\mathbf{X}
+
\phi_{\mathrm{att}}
\left(
\sum_{\ell\in\{1,2\}}
\alpha_{\ell}\mathbf{A}^{(\ell)}
\right).
\end{equation}
Here, $\alpha_{\ell}$ denotes the learnable weight for lag $\ell$, and
$\phi_{\mathrm{att}}(\cdot)$ is a learnable feature transformation.

To emphasize spatial regions with pronounced inter-frame changes, cosine dissimilarity is computed between the enhanced tokens and their shifted counterparts:
\begin{equation}
\mathbf{M}^{(\ell)}
=
1-
\left\langle
\operatorname{Norm}(\mathbf{Y}),
\operatorname{Norm}(\mathbf{Y}^{(\ell)})
\right\rangle .
\end{equation}
After normalization across spatial patches, the resulting weights
$\mathbf{W}^{(\ell)}$ are used to aggregate lag-specific differential features:
\begin{equation}
\mathbf{V}^{(\ell)}
=
\sum_{p=1}^{P}
\mathbf{W}^{(\ell)}_{p}
\left(
\mathbf{Y}^{(\ell)}_{p}-\mathbf{Y}_{p}
\right).
\end{equation}
The lag-specific features are combined with learnable lag embeddings to form compact motion-aware tokens.

The enhanced patch tokens and lag tokens are concatenated with a CLS token and positional embeddings, and are then processed by a Transformer encoder. Multi-head attention pooling with $K$ learnable queries compresses the non-CLS context tokens while retaining multiple anatomical and motion-related summaries. Together with the frame-wise CLS tokens, this yields
\begin{equation}
\mathbf{G}_{\mathrm{us}}
\in
\mathbb{R}^{B\times (K+1)N\times D}.
\end{equation}

\paragraph{Stage 3: temporal encoder.}
A temporal Transformer models dependencies across the full clip:
\begin{equation}
\mathbf{Z}_{\mathrm{us}}
=
\operatorname{Transformer}
\left(
\mathbf{G}_{\mathrm{us}}
\right).
\end{equation}
The output is reshaped into frame-wise groups and pooled over the corresponding $(K+1)$ tokens:
\begin{equation}
\mathbf{H}_{\mathrm{us}}
=
\operatorname{Pool}_{K+1}
\left(
\operatorname{Reshape}(\mathbf{Z}_{\mathrm{us}})
\right)
\in\mathbb{R}^{B\times N\times D}.
\end{equation}
An MLP pose head maps these temporal features to frame-aligned pose predictions:
\begin{equation}
\bar{\boldsymbol{\Theta}}_{\mathrm{us}}
=
g_{\mathrm{pose}}(\mathbf{H}_{\mathrm{us}})
\in\mathbb{R}^{B\times N\times 6}.
\end{equation}
After discarding the first output, the remaining predictions form the relative pose sequence $\Delta \hat{\boldsymbol{\Theta}}_{\mathrm{us}}
\in
\mathbb{R}^{B\times (N-1)\times 6}.
$

During training, the predicted poses are supervised by the multi-scale pose and correlation losses, while $\mathbf{H}_{\mathrm{us}}$ is regularized by the triplet loss. The same temporal features are retained as the anatomical and motion-aware ultrasound representation for cross-modal fusion.

\subsubsection{Global-to-Local Pose Refinement via Dual-Modal Fusion}
\label{subsec:fusion_pipeline}

\begin{figure*}[htp!]
\centering
\includegraphics[width=0.9\textwidth]{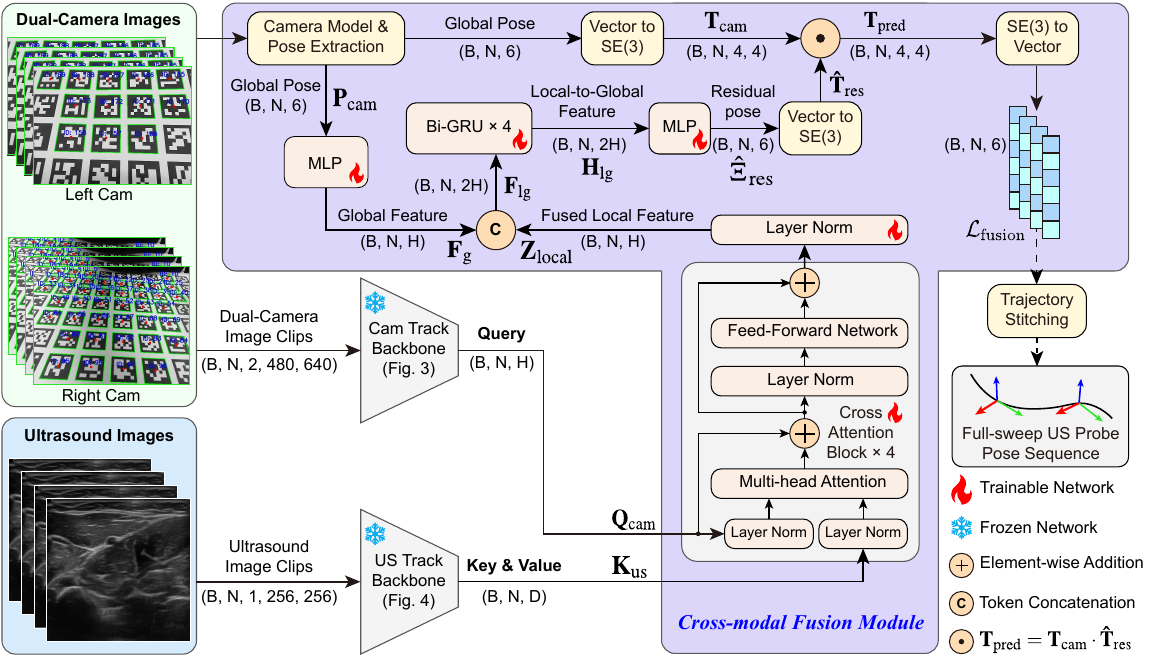}
\caption{Global-to-local pose refinement via cross-modal fusion. Cross-attention integrates frozen camera and ultrasound features to predict residual corrections to the camera-derived global poses. Overlapping-window predictions are stitched to recover the full-sweep probe trajectory.}
\label{fig:fusion}
\end{figure*}

The fusion module refines the camera-derived global probe trajectory using complementary anatomical and motion-related cues from B-mode images. As illustrated in Fig.~\ref{fig:fusion}, the camera branch provides a coarse global pose sequence and contextual query features, while the ultrasound branch provides motion-aware key-value features. The pretrained camera and ultrasound backbones are frozen during this stage, and only the cross-modal fusion module and residual pose head are optimized.

For a window of $N$ synchronized frames, the camera-derived global pose sequence is denoted by
\begin{equation}
\mathbf{P}_{\mathrm{cam}}
=
\left[
\mathbf{p}_{\mathrm{cam},1},
\ldots,
\mathbf{p}_{\mathrm{cam},N}
\right]
\in
\mathbb{R}^{B\times N\times 6},
\end{equation}
where $B$ denotes the batch size. Let
$\mathbf{Q}_{\mathrm{cam}}\in\mathbb{R}^{B\times N\times H}$
and
$\mathbf{K}_{\mathrm{us}}\in\mathbb{R}^{B\times N\times D}$
denote the latent camera and ultrasound features extracted by the frozen backbones, respectively. They are projected into a common hidden space:
\begin{equation}
\mathbf{Q}^{0}
=
\mathbf{Q}_{\mathrm{cam}}\mathbf{W}_{q},
\quad
\mathbf{K}_{v}
=
\mathbf{K}_{\mathrm{us}}\mathbf{W}_{u},
\end{equation}
where
$\mathbf{W}_{q}\in\mathbb{R}^{H\times H}$
and
$\mathbf{W}_{u}\in\mathbb{R}^{D\times H}$.
The projected camera features serve as queries, whereas the ultrasound features serve as keys and values.

A stack of cross-attention blocks progressively injects ultrasound-derived information into the camera representation. In the $\psi$-th block,
\begin{equation}
\bar{\mathbf{Q}}^{\psi}
=
\mathbf{Q}^{\psi-1}
+
\operatorname{MHA}
\left(
\operatorname{LN}(\mathbf{Q}^{\psi-1}),
\operatorname{LN}(\mathbf{K}_{v}),
\operatorname{LN}(\mathbf{K}_{v})
\right),
\end{equation}
followed by
\begin{equation}
\mathbf{Q}^{\psi}
=
\bar{\mathbf{Q}}^{\psi}
+
\operatorname{FFN}
\left(
\operatorname{LN}(\bar{\mathbf{Q}}^{\psi})
\right).
\end{equation}
Here, $\operatorname{LN}(\cdot)$ denotes layer normalization. After $N_{\mathrm{ca}}=4$ blocks, the fused local representation is
\begin{equation}
\mathbf{Z}_{\mathrm{local}}
=
\operatorname{LN}
\left(
\mathbf{Q}^{N_{\mathrm{ca}}}
\right)
\in
\mathbb{R}^{B\times N\times H}.
\end{equation}
This representation retains the globally anchored camera context while selectively incorporating ultrasound cues related to local tissue appearance and probe motion.

The camera-derived global pose sequence is embedded as
\begin{equation}
\mathbf{F}_{g}
=
\phi_{g}
\left(
\mathbf{P}_{\mathrm{cam}}
\right)
\in
\mathbb{R}^{B\times N\times H}.
\end{equation}
It is concatenated with the fused local representation and processed by a Bi-GRU:
\begin{equation}
\mathbf{H}_{\mathrm{lg}}
=
\operatorname{Bi-GRU}
\left(
\operatorname{Concat}
\left(
\mathbf{F}_{g},
\mathbf{Z}_{\mathrm{local}}
\right)
\right)
\in
\mathbb{R}^{B\times N\times 2H}.
\end{equation}
An MLP residual head then predicts frame-wise residual poses:
\begin{equation}
\hat{\boldsymbol{\Xi}}_{\mathrm{res}}
=
\phi_{\mathrm{res}}
\left(
\mathbf{H}_{\mathrm{lg}}
\right)
\in
\mathbb{R}^{B\times N\times 6}.
\end{equation}
The final layer of the residual head is initialized to zero, such that the initial residual transformations are close to identity and the network learns only the corrections supported by the fused observations.

The residual correction is applied in $SE(3)$ rather than through direct addition of pose vectors. For frame $t$, the camera-derived pose
$\mathbf{p}_{\mathrm{cam},t}$
and predicted residual
$\hat{\boldsymbol{\xi}}_{\mathrm{res},t}$
are converted into transformations
$\mathbf{T}_{\mathrm{cam},t}$
and
$\hat{\mathbf{T}}_{\mathrm{res},t}$,
respectively. The refined global pose is obtained by
\begin{equation}
\hat{\mathbf{T}}_{\mathrm{pred},t}
=
\mathbf{T}_{\mathrm{cam},t}
\hat{\mathbf{T}}_{\mathrm{res},t},
\quad
t=1,\ldots,N.
\end{equation}
Thus, the camera trajectory provides the global reference, while ultrasound observations supply internal B-mode cues from anatomical structures, speckle variations, and probe-tissue interaction to correct local errors.

The fusion module is trained using translation and rotation errors on the refined poses, together with a residual regularization term:
\begin{equation}
\mathcal{L}_{\mathrm{fusion}}
=
\mathcal{L}_{\mathrm{trans}}
+
\lambda_{\mathrm{rot}}
\mathcal{L}_{\mathrm{rot}}
+
\lambda_{\mathrm{res}}
\left\|
\hat{\boldsymbol{\Xi}}_{\mathrm{res}}
\right\|_{F}^{2}.
\end{equation}
Here, $\mathcal{L}_{\mathrm{trans}}$ and $\mathcal{L}_{\mathrm{rot}}$ measure the translation and geodesic rotation errors of the refined poses, respectively, as defined in Section~\ref{subsec:loss_functions}. The residual regularization discourages unnecessarily large corrections and preserves the global stability of the camera-derived trajectory.

\subsection{Loss Functions}
\label{subsec:loss_functions}

The camera and ultrasound tracking backbones are trained using three complementary objectives: a multi-scale pose loss for geometric accuracy, a correlation loss for sequence-level motion consistency, and a triplet loss for motion-discriminative feature learning.

\subsubsection{Multi-Scale Pose Loss}

Let $\Delta\hat{\boldsymbol{\theta}}_i$ and
$\Delta\boldsymbol{\theta}_i\in\mathbb{R}^{6}$ denote the predicted and ground-truth relative poses between frames $i$ and $i+1$, respectively. They are converted into homogeneous transformations
$\Delta\hat{\mathbf{T}}_i,\Delta\mathbf{T}_i\in SE(3)$.
Rather than supervising only adjacent-frame motion, we evaluate composed transformations over multiple temporal horizons to reduce accumulated trajectory drift.

For a horizon $k$, the ground-truth and predicted transformations are
\begin{equation}
\Delta\mathbf{T}_{i}^{(k)}
=
\prod_{r=i}^{i+k-1}\Delta\mathbf{T}_{r},
\quad
\Delta\hat{\mathbf{T}}_{i}^{(k)}
=
\prod_{r=i}^{i+k-1}\Delta\hat{\mathbf{T}}_{r},
\end{equation}
where the products follow the temporal order of the scan. Their translation and rotation components are denoted by
$(\mathbf{t}_{i}^{(k)},\mathbf{R}_{i}^{(k)})$
and
$(\hat{\mathbf{t}}_{i}^{(k)},\hat{\mathbf{R}}_{i}^{(k)})$,
respectively.

The translation loss at horizon $k$ is
\begin{equation}
\mathcal{L}_{\mathrm{trans}}^{(k)}
=
\frac{1}{|\mathcal{I}_k|}
\sum_{i\in\mathcal{I}_k}
\rho_{\delta}
\left(
\left\|
\hat{\mathbf{t}}_{i}^{(k)}
-
\mathbf{t}_{i}^{(k)}
\right\|_2
\right),
\end{equation}
where $\mathcal{I}_k$ is the set of valid starting indices for $k$-step composition and
\begin{equation}
\rho_{\delta}(r)
=
\begin{cases}
\frac{1}{2}r^2, & r\leq\delta,\\
\delta\left(r-\frac{1}{2}\delta\right), & r>\delta.
\end{cases}
\end{equation}

The rotation loss is defined using the geodesic distance on $SO(3)$:
\begin{equation}
\mathcal{L}_{\mathrm{rot}}^{(k)}
=
\frac{1}{|\mathcal{I}_k|}
\sum_{i\in\mathcal{I}_k}
d_R
\left(
\mathbf{R}_{i}^{(k)},
\hat{\mathbf{R}}_{i}^{(k)}
\right).
\end{equation}
Let
\begin{equation}
\mathbf{R}_{e,i}^{(k)}
=
\left(\mathbf{R}_{i}^{(k)}\right)^{\top}
\hat{\mathbf{R}}_{i}^{(k)}.
\end{equation}
The geodesic distance is computed as
\begin{equation}
d_R
\left(
\mathbf{R}_{i}^{(k)},
\hat{\mathbf{R}}_{i}^{(k)}
\right)
=
\arctan2
\left(
a_i^{(k)},
b_i^{(k)}
\right),
\end{equation}
where
\begin{equation}
a_i^{(k)}
=
\frac{1}{2}
\left\|
\left(
\mathbf{R}_{e,i}^{(k)}
-
{\mathbf{R}_{e,i}^{(k)}}^{\top}
\right)^{\vee}
\right\|_2,
\quad
b_i^{(k)}
=
\frac{
\operatorname{tr}(\mathbf{R}_{e,i}^{(k)})-1
}{2}.
\end{equation}
Here, $(\cdot)^{\vee}$ maps a skew-symmetric matrix in
$\mathfrak{so}(3)$ to its corresponding vector in
$\mathbb{R}^{3}$.

Translation and rotation losses are balanced using uncertainty-based weighting~\citep{kendall2018multi}:
\begin{equation}
\mathcal{L}_{\mathrm{pose}}^{(k)}
=
\exp(-s_t)\mathcal{L}_{\mathrm{trans}}^{(k)}
+
s_t
+
\exp(-s_r)\mathcal{L}_{\mathrm{rot}}^{(k)}
+
s_r,
\end{equation}
where $s_t$ and $s_r$ are learnable log-uncertainty parameters.

The final multi-scale pose loss is
\begin{equation}
\mathcal{L}_{\mathrm{pose}}
=
\frac{
\sum_{k\in\mathcal{K}}
w_k\mathcal{L}_{\mathrm{pose}}^{(k)}
}{
\sum_{k\in\mathcal{K}}w_k
}.
\end{equation}
We set $w_k=1/\sqrt{k}$ to moderately down-weight longer horizons while retaining sensitivity to systematic drift. Further discussion of this weighting strategy is provided in Appendix \ref{supp:multiscale_weighting}.

\subsubsection{Correlation Loss}

Although the multi-scale pose loss constrains geometric error, it does not directly encourage the predicted sequence to follow the overall motion pattern of the target. We therefore apply a cosine correlation loss to the normalized relative pose sequences.

Let
$\Delta\hat{\boldsymbol{\Theta}},
\Delta\boldsymbol{\Theta}
\in\mathbb{R}^{B\times(N-1)\times6}$
denote the predicted and ground-truth sequences, and let
$\hat{\mathbf{v}}
=
\operatorname{vec}
\left(
\Delta\hat{\boldsymbol{\Theta}}
\right),
\mathbf{v}
=
\operatorname{vec}
\left(
\Delta\boldsymbol{\Theta}
\right).
$
The loss is defined as
\begin{equation}
\mathcal{L}_{\mathrm{corr}}
=
1
-
\frac{
\left\langle
\hat{\mathbf{v}},
\mathbf{v}
\right\rangle
}{
\left\|
\hat{\mathbf{v}}
\right\|_2
\left\|
\mathbf{v}
\right\|_2
}.
\end{equation}
This objective encourages agreement between the overall directions of the predicted and ground-truth motion sequences.

\subsubsection{Triplet Loss}

A triplet loss is applied to the temporal latent features to encourage motion-discriminative representations. Let
$\mathbf{H}_b\in\mathbb{R}^{N\times D_h}$
denote the latent feature sequence of sample $b$, and let
$\mathbf{y}_b\in\mathbb{R}^{(N-1)\times6}$
denote its normalized target pose sequence. Their vectorized forms are
\begin{equation}
\mathbf{h}_b
=
\operatorname{vec}(\mathbf{H}_b),
\quad
\tilde{\mathbf{y}}_b
=
\operatorname{vec}(\mathbf{y}_b).
\end{equation}

For an anchor $a$, a positive sample $p$, and a negative sample $n$, triplets are selected such that
\begin{equation}
d_y
\left(
\tilde{\mathbf{y}}_a,
\tilde{\mathbf{y}}_p
\right)
<
d_y
\left(
\tilde{\mathbf{y}}_a,
\tilde{\mathbf{y}}_n
\right),
\end{equation}
where $d_y(\cdot,\cdot)$ measures distance in the normalized pose space. The triplet loss is
\begin{equation}
\mathcal{L}_{\mathrm{triplet}}
=
\frac{1}{|\mathcal{Q}|}
\sum_{(a,p,n)\in\mathcal{Q}}
\max
\left(
0,
d_h(\mathbf{h}_a,\mathbf{h}_p)
-
d_h(\mathbf{h}_a,\mathbf{h}_n)
+
m
\right),
\end{equation}
where $\mathcal{Q}$ is the set of sampled triplets,
$m$ is the margin, and $d_h(\cdot,\cdot)$ is the Euclidean distance in the latent feature space.

The total loss for training the camera and ultrasound backbones is
\begin{equation}
\mathcal{L}_{\mathrm{total}}
=
\mathcal{L}_{\mathrm{pose}}
+
\lambda_{\mathrm{corr}}
\mathcal{L}_{\mathrm{corr}}
+
\lambda_{\mathrm{triplet}}
\mathcal{L}_{\mathrm{triplet}},
\end{equation}
where $\lambda_{\mathrm{corr}}$ and
$\lambda_{\mathrm{triplet}}$ control the contributions of the sequence-consistency and feature-regularization terms, respectively.

\section{Experiments and Evaluation}
\label{sec:experiments}

In this section, we first describe the datasets, acquisition protocols, implementation details, and evaluation metrics used in the experiments. We then present quantitative and qualitative results on FUSION-J, FUSION-L, and the public TUS-REC2025 dataset to assess pose estimation accuracy, trajectory consistency, and 3-D reconstruction quality.

\subsection{Experimental Setup}
\label{subsec:exp_setup}

\begin{figure*}[htp!]
\centering
\includegraphics[width=0.95\textwidth]{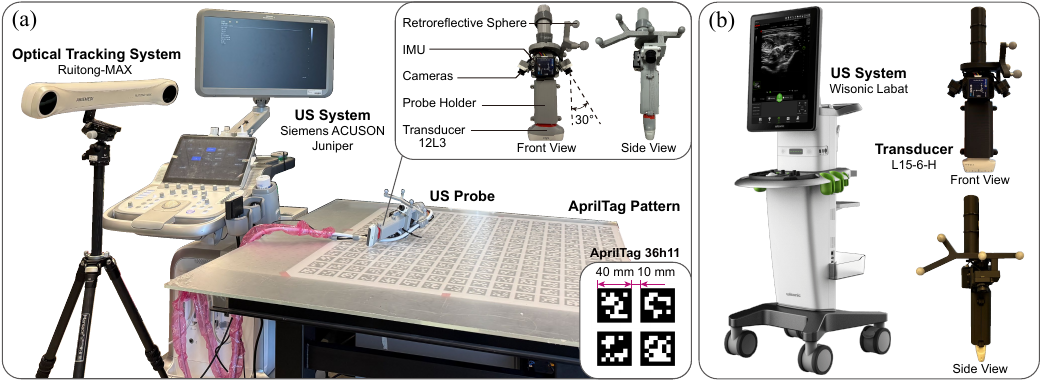}
\caption{Overview of the data acquisition system. B-mode ultrasound frames were acquired using a clinical ultrasound machine, while dual cameras mounted on the probe holder captured the AprilTag pattern for camera-based pose estimation. An optical tracking system was used to record ground-truth probe poses.}
\label{fig:setup}
\end{figure*}

All in-house experiments used a common hardware platform illustrated in Fig.~\ref{fig:setup}, consisting of three subsystems: a visual pose estimation subsystem, a ground-truth recording subsystem, and an ultrasound imaging subsystem.

\textit{Visual pose estimation.} Two low-cost monocular RGB cameras (2303U, China, \texteuro15 each) were rigidly mounted on a custom 3-D-printed probe holder to observe the scanning scene from complementary viewpoints. Each camera captured $480\times640$ RGB images at 30~fps with an $83^\circ$ horizontal field of view, and camera intrinsic parameters were calibrated to support AprilTag-based pose estimation. The visual reference was a planar AprilTag board containing a dense $15\times15$ grid of markers, each measuring $40~\mathrm{mm}\times40~\mathrm{mm}$ with a $10~\mathrm{mm}$ separation along both axes. This dense layout provides multiple fiducial correspondences per camera view, improving pose estimation robustness.

\textit{Ground-truth recording.} An optical tracking system (Ruitong-MAX, ARIEMEDI, China) recorded the 6-DoF probe pose at 120~fps via a rigid marker co-located with the probe holder. These optical measurements served exclusively as ground-truth annotations for training and evaluation, and were not provided as input to the proposed vision-based model.

\textit{Ultrasound imaging.} Two different ultrasound systems were used across the two in-house datasets: an ACUSON Juniper machine (Siemens AG, Germany) with a 12L3 linear array probe and a frame grabber (CM410, UGREEN, China) capturing B-mode images at 30~fps, and a Wisonic Labat system (Wisonic, China) with an L15-6-H linear array probe capturing at 10~fps.

All three subsystems were operated simultaneously and temporally aligned using ROS2 on Ubuntu~22.04. The synchronized dataset therefore contains paired US images, AprilTag-based camera pose estimates, and optical-tracking ground-truth poses for supervised training and quantitative evaluation.

\subsection{Data Preparation}
\label{subsec:data_preparation}

\subsubsection{FUSION-J}
\label{subsubsec:fusion_j}

The Freehand Ultrasound and viSION dataset -- Juniper (FUSION-J) was collected using an ACUSON Juniper ultrasound machine. Data were collected from eight participants, with scans performed on both the left and right forearms. For each forearm, four scanning trajectories were acquired: C-shaped, linear, S-shaped, and zigzag. Each trajectory was repeated three times, resulting in 192 scans in total. After excluding scans with poor image quality or tracking errors, 164 scans remained, with approximately 1000 frames per scan. The dataset was split at the scan level into training, validation, and test sets containing 132, 16, and 16 scans, respectively, corresponding to 80.48\%, 9.76\%, and 9.76\% of the final dataset. The dataset will be publicly released.

\subsubsection{FUSION-L}
\label{subsubsec:fusion_l}

The Freehand Ultrasound and viSION dataset -- Labat (FUSION-L) was collected to evaluate the robustness of the proposed model across different ultrasound hardware configurations. The same acquisition setup was used, with the ultrasound system replaced by the Wisonic Labat machine. Data were collected from ten participants, with scans performed on both the left and right forearms. For each forearm, the same four trajectory patterns were acquired, each repeated five times, resulting in 400 scans in total. After excluding scans with poor image quality or tracking errors, 391 scans remained, with approximately 300 frames per scan. The dataset was split into training, validation, and test sets containing 311, 40, and 40 scans, respectively, corresponding to 79.54\%, 10.23\%, and 10.23\% of the final dataset. The dataset will be publicly released.

\subsubsection{TUS-REC2025}
\label{subsubsec:tus_rec}

In addition to the two in-house datasets, we evaluated the proposed model on the public freehand ultrasound dataset TUS-REC2025\footnote{\url{https://zenodo.org/records/15699959}}. The dataset includes left and right forearm scans acquired using fanning and rocking motions around a fixed contact point near the elbow, repeated after rotating the probe by approximately $90^\circ$. It contains 106 scans with around 1600 frames per scan, split into training, validation, and test sets of 86, 10, and 10 scans, respectively, corresponding to 81.14\%, 9.43\%, and 9.43\% of the dataset. Only B-mode images and tracking poses are available; this dataset was therefore used for the ultrasound-only branch only.

\subsection{Implementation Details}
\label{subsec:implementation_details}

For each scan sequence, B-mode US frames, dual-camera RGB images, and tracking poses were temporally synchronized before training. The US frames were converted to single-channel tensors, normalized to $[0,1]$, center-cropped, and resized to $256 \times 256$. The ground-truth relative transformations were computed from the synchronized tracking poses following the definition in Section~\ref{subsec:problem_definition}.

Full scan sequences were divided into overlapping 16-frame windows with a stride of one, yielding 15 relative transformations per window. The same configuration was used for the ultrasound-only, camera-only, and dual-modal models. The ultrasound and camera branches were first trained independently. For dual-modal residual refinement, the modality-specific encoders were initialized from the independently trained models and kept frozen, while the fusion module was optimized to estimate residual corrections to the camera-derived poses. The multi-scale pose loss used temporal horizons $\mathcal{K}=\{1,2,4,8,15\}$ to supervise both adjacent-frame motion and composed motion over longer intervals.

Random sweep reversal and temporally consistent patch cutout were applied during training, each with a probability of $0.5$. For patch cutout, $20\%$ of the $16\times16$ image patches were masked using a common mask across all frames in a window, thereby avoiding artificial temporal changes inconsistent with the pose labels.

All models were implemented in PyTorch and trained on four NVIDIA GeForce RTX 3090 GPUs with an effective batch size of 96. AdamW was used with a base learning rate of $1\times10^{-4}$ and a weight decay of $1\times10^{-4}$. The models were trained for 20 epochs using one warm-up epoch followed by cosine learning-rate decay.

During inference, each full scan was divided into the same overlapping 16-frame windows used during training. Consequently, each frame-to-frame motion could be estimated from multiple windows. These estimates were combined using triangular weights, which assign greater importance to predictions near the center of a window and lower importance to those near its boundaries. The resulting relative motions were then composed to recover a continuous probe trajectory for the full scan.

We evaluated frame-wise pose accuracy using the mean absolute error (MAE) of the six global pose components. For trajectory-level consistency, predicted relative transformations were accumulated and aligned to the first reference frame, and we reported the final drift rate (FDR), average drift rate (ADR), maximum drift (MD), average drift (AD), and trajectory Hausdorff distance ($\mathrm{HD}_{\mathrm{traj}}$). Model size is reported in MB according to the number of trainable parameters. Formal definitions of the pose, trajectory, and reconstruction metrics are provided in Appendix \ref{supp:reconstruction_metrics}.

\subsection{Estimation Results and Ablation Studies}
\label{subsec:results}
\subsubsection{Ultrasound-only Pose Estimation Results}
\label{subsubsec:ultrasound_only_results}

Table~\ref{tab:dataset1} reports the quantitative results of the ultrasound-only model on the two in-house datasets (FUSION-J and FUSION-L) and the public TUS-REC2025 dataset. The proposed model was compared with representative learning-based methods using the same evaluation protocol.

\begin{table*}[t]
\centering
\caption{Ultrasound-only pose estimation performance on three datasets. Baseline methods include CNN~\citep{prevost20183d}, 
DC$^2$-Net~\citep{guo2020sensorless,guo2022ultrasound}, 
USFormer~\citep{ning2022spatial}, 
Li et al.~\citep{li2023long}, 
FiMA~\citep{yan2024fine}, 
MoNet~\citep{luo2022deep,luo2025monetv2}, 
MoGLo~\citep{lee2025enhancing}, and 
DualTrack~\citep{wilson2025dualtrack}. 
The best and second-best results in each dataset are highlighted in bold and underlined, respectively.}
\label{tab:dataset1}
\scriptsize
\setlength{\tabcolsep}{3.2pt}
\renewcommand{\arraystretch}{1}
\resizebox{\textwidth}{!}{
\begin{tabular}{clcccccccccccc}
\toprule
\multirow{2}{*}{Dataset}
& \multirow{2}{*}{Model}
& \multicolumn{6}{c}{Pose MAE (mm/$^{\circ}$) $\downarrow$}
& \multicolumn{5}{c}{Trajectory Drift Metrics (\% / mm) $\downarrow$}
& \multirow{2}{*}{\shortstack{Params\\(MB)}}\\
\cmidrule(lr){3-8} \cmidrule(lr){9-13}
&
& X
& Y
& Z
& Roll
& Pitch
& Yaw 
& FDR
& ADR
& MD
& AD
& $\mathrm{HD}_{\mathrm{traj}}$
&
\\
\midrule

\multirow{9}{*}{FUSION-J}
& CNN
& 13.05 & 7.23 & 15.56 & 4.94 & 7.39 & 5.63
& 5.03\% & 14.84\% & 48.35 & 24.59 & 35.13 & 23.51 \\

& DC$^2$-Net
& \textbf{6.93} & \underline{5.71} & 11.81 & 3.05 & 4.41 & 4.87 
& 3.49\% & 7.30\% & 33.84 & 16.56 & 25.08 & 27.92 \\

& USFormer
& 11.27 & 5.77 & \underline{9.96} & \textbf{2.96} & 4.81 & 8.48 
& 3.29\% & 7.53\% & 32.93 & 18.05 & 27.37 & 28.69 \\

& Li et al.
& 12.41 & 6.33 & 12.89 & 3.82 & 5.76 & 6.46 
& 3.60\% & 10.11\% & 40.91 & 21.80 & 33.32 & 26.98 \\

& FiMA
& 8.66 & 7.28 & 14.83 & 3.54 & 4.54 & 5.01 
& 4.40\% & 8.58\% & 42.17 & 21.81 & 32.19 & 26.35 \\

& MoNet
& 7.68 & 6.07 & 16.29 & 3.57 & 5.12 & 4.76 
& 3.45\% & 9.30\% & 41.03 & 21.44 & 30.39 & 30.04 \\

& MoGLo
& 8.24 & 7.03 & 10.31 & 3.15 & \underline{4.27} & \textbf{3.94} 
& \underline{3.26\%} & {6.78\%} & \underline{32.65} & 17.52 & \textbf{23.98} & 27.90 \\

& DualTrack
& \underline{7.22} & \textbf{5.33} & 11.86 & 3.13 & \textbf{4.05} & 4.32 
& 3.53\% & \underline{6.27\%} & 33.33 & \underline{16.39} & 24.60 & 23.68 \\

& Ours
& 7.76 & 6.08 & \textbf{8.83} & \underline{3.00} & 4.36 & \underline{4.26} 
& \textbf{3.17\%} & \textbf{5.80\%} & \textbf{30.07} & \textbf{15.14} & \underline{24.52} & 26.01 \\

\midrule

\multirow{9}{*}{FUSION-L}
& CNN
& 14.07 & 4.68 & 11.04 & 3.76 & 12.56 & 4.37
& 4.80\% & 9.04\% & 45.16 & 21.10 & 35.89 & 23.51 \\

& DC$^2$-Net
& 11.17 & 4.02 & 7.78 & 3.89 & 9.70 & 4.48
& 3.74\% & 7.09\% & 33.52 & 15.86 & 27.85 & 27.92 \\

& USFormer
& \underline{9.91} & 4.29 & 7.89 & \textbf{2.81} & 8.58 & 4.21
& 3.82\% & 6.78\% & 33.00 & 15.25 & 28.02 & 28.69 \\

& Li et al.
& 15.50 & 4.75 & 10.49 & 4.70 & 13.76 & 7.21
& 5.20\% & 9.99\% & 47.78 & 21.57 & 39.41 & 26.98 \\

& FiMA
& 10.12 & \underline{3.73} & 8.03 & 3.36 & 8.36 & \textbf{3.92}
& 3.58\% & 6.60\% & 32.90 & 15.17 & 27.09 & 26.35 \\

& MoNet
& 12.43 & 4.43 & 10.34 & 3.60 & 11.19 & 5.46
& 4.45\% & 8.30\% & 39.82 & 18.94 & 31.79 & 30.04 \\

& MoGLo
& 10.51 & 3.85 & 7.30 & 3.63 & 9.25 & 4.49
& 3.49\% & \underline{6.19\%} & 32.38 & 15.14 & 27.47 & 27.90 \\

& DualTrack
& 10.05 & 3.85 & \underline{7.12} & \underline{2.94} & \textbf{7.55} & \underline{3.94}
& \underline{3.21\%} & 6.35\% & \underline{30.38} & \underline{14.50} & \underline{25.79} & 23.68 \\

& Ours
& \textbf{9.46} & \textbf{3.43} & \textbf{6.74} & 3.26 & \underline{8.00} & 4.62
& \textbf{3.09\%} & \textbf{5.95\%} & \textbf{29.52} & \textbf{13.51} & \textbf{24.84} & 26.01 \\

\midrule

\multirow{9}{*}{TUS-REC2025}
& CNN
& 6.31 & \underline{2.87} & 9.07 & 17.17 & 13.48 & 8.25
& 3.87\% & 13.66\% & 24.85 & 12.67 & 20.98 & 23.51 \\

& DC$^2$-Net
& \underline{2.55} & 6.98 & 9.25 & \textbf{10.20} & \underline{4.43} & \textbf{5.49}
& 4.31\% & 12.07\% & 20.65 & 12.67 & 17.36 & 27.92 \\

& USFormer
& 2.88 & 4.92 & \underline{8.81} & 14.86 & 9.51 & \underline{6.73}
& \underline{3.34\%} & 11.06\% & 20.44 & 11.56 & 16.48 & 28.69 \\

& Li et al.
& 10.99 & \textbf{2.83} & 9.83 & \underline{11.74} & 24.50 & 8.72
& 5.55\% & 17.16\% & 36.85 & 16.85 & 28.07 & 26.98 \\

& FiMA
& 2.65 & 4.50 & 8.82 & 14.62 & 5.01 & 8.06
& 3.45\% & \underline{10.94\%} & \textbf{18.12} & \underline{11.19} & \underline{15.38} & 26.35 \\

& MoNet
& 10.93 & 4.08 & 10.06 & 12.98 & 23.00 & 7.53
& 6.16\% & 17.18\% & 36.17 & 17.45 & 27.78 & 30.04 \\

& MoGLo
& 3.62 & 5.19 & 8.91 & 17.72 & 6.43 & 9.10
& 4.18\% & 12.59\% & 20.91 & 12.03 & 18.45 & 27.90 \\

& DualTrack
& \textbf{1.66} & 5.12 & 11.92 & 13.73 & \textbf{3.25} & 8.32
& 4.54\% & 11.35\% & 20.89 & 14.21 & 18.91 & 23.68 \\

& Ours
& \underline{2.55} & 3.83 & \textbf{8.09} & 15.47 & {4.81} & 8.89
& \textbf{2.93\%} & \textbf{9.69\%} & \underline{18.29} & \textbf{10.37} & \textbf{14.90} & 26.01 \\

\bottomrule
\end{tabular}
}
\end{table*}

On FUSION-J, the proposed ultrasound-only model achieved the best $Z$-translation accuracy and competitive rotational accuracy. More importantly, it obtained the lowest FDR, ADR, MD, and AD, despite not achieving the lowest error for every individual pose component. In particular, it reduced ADR to 5.80\% and AD to 15.14~mm, demonstrating improved long-term trajectory consistency.

On FUSION-L, the proposed model achieved the lowest translational MAE along all three axes and the best performance across all trajectory-level metrics. It reduced AD from the previous best value of 14.50~mm to 13.51~mm and $\mathrm{HD}_{\mathrm{traj}}$ from 25.79~mm to 24.84~mm, indicating consistent improvements in both frame-wise accuracy and accumulated trajectory estimation.

On TUS-REC2025, the proposed model achieved the lowest FDR, ADR, AD, and $\mathrm{HD}_{\mathrm{traj}}$, while remaining competitive in the frame-wise pose metrics. Specifically, ADR decreased from 10.94\% to 9.69\%, and $\mathrm{HD}_{\mathrm{traj}}$ decreased from 15.38~mm to 14.90~mm. Its MD was 18.29~mm, close to the best value of 18.12~mm.

Overall, these results show that the proposed motion-aware ultrasound branch primarily improves long-term trajectory consistency rather than uniformly minimizing every frame-wise pose component. Its consistent performance across the three datasets provides a strong foundation for subsequent dual-modal residual refinement.

\subsubsection{Camera-only Pose Estimation Results}
\label{subsubsec:camera_only}

Table~\ref{tab:ablation_modalities} reports the camera-based pose estimation results on FUSION-J and FUSION-L. Even a single camera substantially outperformed the ultrasound-only model in both frame-wise pose accuracy and trajectory-level consistency, confirming that external visual observations provide a stable spatial reference for freehand probe motion estimation.

\begin{table*}[t]
\centering
\caption{Pose estimation performance of camera-only, ultrasound-only, and dual-modal models on FUSION-J and FUSION-L. The best and second-best results in each dataset are highlighted in bold and underlined, respectively.}
\label{tab:ablation_modalities}
\scriptsize
\setlength{\tabcolsep}{3.2pt}
\renewcommand{\arraystretch}{1}
\resizebox{\textwidth}{!}{
\begin{tabular}{clcccccccccccc}
\toprule
\multirow{2}{*}{Dataset}
& \multirow{2}{*}{Input Modality}
& \multicolumn{6}{c}{Pose MAE (mm / $^{\circ}$) $\downarrow$}
& \multicolumn{5}{c}{Trajectory Drift Metrics (\% / mm) $\downarrow$}
& \multirow{2}{*}{\shortstack{Params\\(MB)}} \\
\cmidrule(lr){3-8} \cmidrule(lr){9-13}
&
& X
& Y
& Z
& Roll
& Pitch
& Yaw
& FDR
& ADR
& MD
& AD
& $\mathrm{HD}_{\mathrm{traj}}$
& \\
\midrule

\multirow{8}{*}{FUSION-J}
& L-Cam (Global)
& 0.94 & 1.11 & 1.43 & 0.41 & 0.30 & 0.42
& 0.18\% & 2.48\% & 8.25 & 2.35 & 4.88 & -- \\

& R-Cam (Global)
& 0.88 & 1.26 & \underline{1.21} & 0.38 & \underline{0.26} & 0.46 
& 0.22\% & 2.27\% & 7.32 & 2.27 & 4.88 & -- \\

& Dual-Cam (Global)
& \underline{0.85} & \underline{0.88} & 1.22 & \underline{0.36} & \textbf{0.19} & \underline{0.39} 
& \underline{0.17\%} & 1.89\% & \underline{6.99} & \underline{2.00} & \underline{4.35} & -- \\

& L-Cam (Local)
& 1.89 & 2.07 & 1.85 & 0.55 & 0.55 & 0.99
& 0.80\% & 2.18\% & 9.55 & 3.80 & 6.90 & 1.12 \\

& R-Cam (Local)
& 1.33 & 1.71 & 1.38 & 0.44 & 0.57 & 0.61 
& 0.54\% & \underline{1.84\%} & 8.52 & 2.98 & 5.60 & 1.12 \\

& Dual-Cam (Local)
& 1.09 & 1.04 & 1.71 & 0.42 & 0.48 & 0.41 
& 0.48\% & \textbf{1.59\%} & 7.42 & 2.59 & 4.63 & 1.91 \\

& US only (Local)
& 7.76 & 6.08 & 8.83 & 3.00 & 4.36 & 4.26
& 3.17\% & 5.80\% & 30.07 & 15.14 & 24.52 & 26.01 \\

& US + Dual-Cam
& \textbf{0.71} & \textbf{0.69} & \textbf{1.06} & \textbf{0.29} & \textbf{0.19} & \textbf{0.31} 
& \textbf{0.14\%} & 1.85\% & \textbf{6.61} & \textbf{1.67} & \textbf{4.00} & 4.06 \\

\midrule

\multirow{8}{*}{FUSION-L}
& L-Cam (Global)
& 1.23 & 0.72 & 1.48 & 0.26 & 0.37 & 0.31
& 0.17\% & 2.16\% & 6.12 & 2.34 & 4.09 & -- \\

& R-Cam (Global)
& \underline{0.89} & 1.37 & 1.17 & 0.43 & 0.36 & 0.38 
& 0.17\% & 1.95\% & 5.83 & 2.30 & 4.61 & -- \\

& Dual-Cam (Global)
& 0.95 & \underline{0.71} & \underline{0.97} & \underline{0.22} & \textbf{0.29} & \underline{0.28} 
& \underline{0.13\%} & \underline{1.53\%} & \underline{4.34} & \underline{1.77} & \underline{3.37} & -- \\

& L-Cam (Local)
& 1.35 & 1.28 & 1.89 & 0.43 & 0.80 & 0.62
& 0.61\% & 1.99\% & 6.95 & 3.05 & 5.14 & 1.12 \\

& R-Cam (Local)
& 1.32 & 1.07 & 1.67 & 0.42 & 0.67 & 0.50 
& 0.55\% & 1.72\% & 6.41 & 2.75 & 5.02 & 1.12 \\

& Dual-Cam (Local)
& 1.31 & 0.82 & 1.67 & 0.63 & 0.69 & 0.58 
& 0.46\% & 1.73\% & 5.88 & 2.63 & 4.54 & 1.91 \\

& US only (Local)
& 9.46 & 3.43 & 6.74 & 3.26 & 8.00 & 4.62
& 3.09\% & 5.95\% & 29.52 & 13.51 & 24.84 & 26.01 \\

& US + Dual-Cam
& \textbf{0.62} & \textbf{0.36} & \textbf{0.87} & \textbf{0.14} & \underline{0.30} & \textbf{0.24} 
& \textbf{0.12\%} & \textbf{1.26\%} & \textbf{4.04} & \textbf{1.29} & \textbf{2.96} & 4.06 \\

\bottomrule
\end{tabular}
}
\end{table*}

The table includes two camera-only settings, denoted as Global and Local. In the Global setting, the camera pose is estimated relative to the AprilTag board reference frame and then transformed into the trajectory of the US image center using the refined hand-eye calibration. Because all poses are recovered in a common world coordinate system, this setting avoids the error accumulation caused by sequentially composing local pose estimates, thereby providing more accurate camera-based trajectory estimates.

The Local setting follows the camera-branch formulation described in Section~\ref{subsec:camera_pipeline}. Each complete camera pose sequence is divided into overlapping 16-frame windows, within which the network estimates frame-to-frame relative probe motion. The predictions from these windows are then combined to recover the full trajectory. This setting uses the same window-based relative-motion estimation as the ultrasound model, but the limited temporal context of each window and the subsequent trajectory stitching introduce additional accumulated error relative to the Global setting.

Compared with either single-camera configuration, Dual-Cam generally improved pose accuracy and trajectory consistency in both settings. For example, in the Global setting, it reduced AD to 2.00~mm on FUSION-J and 1.77~mm on FUSION-L. More importantly, the two views provide complementary visual coverage and redundancy: when one view becomes unreliable because of partial occlusion, motion blur, or an unfavorable viewpoint, the other can still provide usable scene observations. The multi-view configuration can therefore reduce dependence on any single camera and support more stable pose estimation under view-specific degradation.

\subsubsection{Dual-Modal Fusion and Ablation Analysis}
\label{subsubsec:multimodal_fusion}

Table~\ref{tab:ablation_modalities} summarizes the pose estimation performance of the proposed US~+~Dual-Cam model and provides a modality ablation through comparison with its US-only and Dual-Cam counterparts. The fusion model achieved the best frame-wise pose accuracy on most components and the lowest FDR, MD, AD, and $\mathrm{HD}_{\mathrm{traj}}$ on both FUSION-J and FUSION-L, demonstrating accurate and globally consistent trajectory estimation across different ultrasound systems.

The comparison with US only isolates the contribution of external visual guidance. Adding dual-camera observations markedly reduced both frame-wise pose errors and accumulated trajectory drift, showing that camera-derived poses provide the global spatial reference that cannot be reliably recovered from B-mode images alone. The comparison with Dual-Cam then isolates the contribution of ultrasound information. Although the camera-only baseline was already strong, incorporating B-mode features further improved most pose and trajectory metrics. In particular, AD decreased from 2.00~mm to 1.67~mm on FUSION-J and from 1.77~mm to 1.29~mm on FUSION-L, corresponding to reductions of 16.50\% and 27.12\%, respectively.

These ablation results demonstrate that the two modalities play complementary roles rather than providing redundant estimates. Their integration preserves the global stability of visual pose estimation while using ultrasound evidence to refine local alignment, supporting the proposed global-to-local residual refinement design.

\subsubsection{Representative Freehand Sweep Evaluation}

\begin{figure*}[htp!]
\centering
\includegraphics[width=0.85\textwidth]{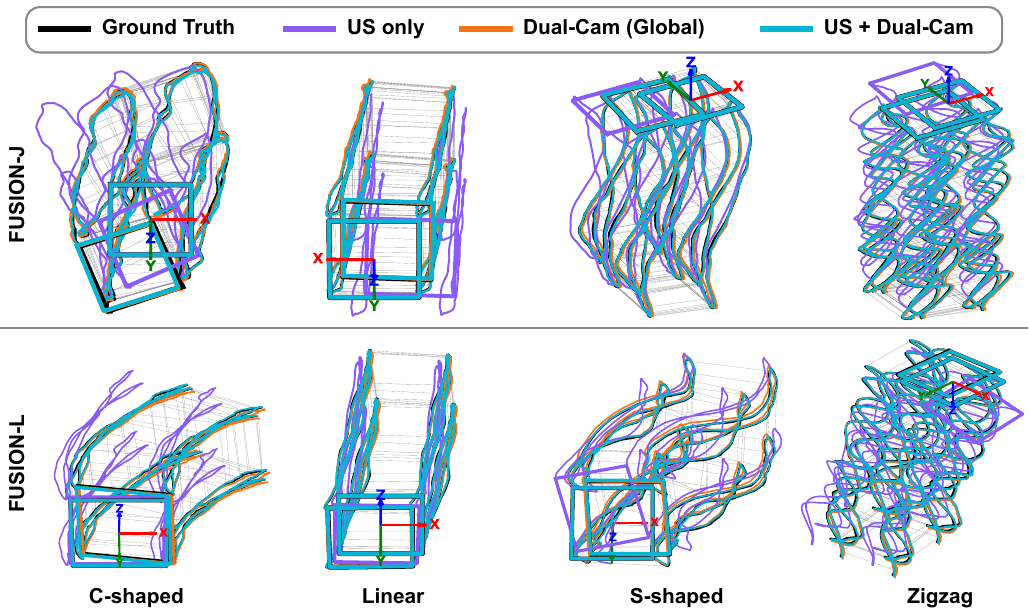}
\caption{Qualitative trajectory comparisons on representative examples from FUSION-J and FUSION-L under four sweep patterns.}
\label{fig:corner_traj}
\end{figure*}

To assess performance across representative freehand sweeps, we evaluated the three models on \textit{in vivo} forearm scans following C-shaped, linear, S-shaped, and zigzag trajectories. As shown in Fig.~\ref{fig:corner_traj}, the US-only trajectories deviate substantially from the reference, particularly during direction changes, whereas Dual-Cam provides a stable global estimate. The proposed US~+~Dual-Cam model preserves this global consistency while improving local trajectory agreement in most examples.

\begin{table*}[t]
\centering
\caption{Quantitative trajectory comparisons on representative examples from FUSION-J and FUSION-L under four sweep patterns. The best result for each sweep pattern is highlighted in bold.}
\label{tab:single_traj_comp}
\scriptsize
\setlength{\tabcolsep}{3.2pt}
\renewcommand{\arraystretch}{1}
\resizebox{\textwidth}{!}{
\begin{tabular}{cllccccccccccc}
\toprule
\multirow{2}{*}{Dataset}
& \multirow{2}{*}{Trajectory}
& \multirow{2}{*}{Method}
& \multicolumn{6}{c}{Pose MAE (mm / $^{\circ}$) $\downarrow$}
& \multicolumn{5}{c}{Trajectory Drift Metrics (\% / mm) $\downarrow$}
\\
\cmidrule(lr){4-9} \cmidrule(lr){10-14}
&
&
& X
& Y
& Z
& Roll
& Pitch
& Yaw
& FDR
& ADR
& MD
& AD
& $\mathrm{HD}_{\mathrm{traj}}$
\\
\midrule

\multirow{12}{*}{FUSION-J}
& \multirow{3}{*}{C-shaped}
& US only
& 11.68 & 3.54 & 4.24 & 2.33 & 9.66 & 2.95
& 3.16\% & 4.86\% & 37.24 & 14.35 & 34.44 \\

&
& Dual-Cam (Global)
& 0.72 & 0.90 & 2.65 & \textbf{0.21} & 0.25 & 0.67
& \textbf{0.30\%} & 1.86\% & 11.34 & 3.11 & 4.25 \\

&
& US + Dual-Cam
& \textbf{0.58} & \textbf{0.69} & \textbf{1.28} & 0.40 & \textbf{0.20} & \textbf{0.35}
& 0.33\% & \textbf{1.47\%} & \textbf{5.70} & \textbf{1.75} & \textbf{3.27} \\

& \multirow{3}{*}{Linear}
& US only
& 5.99 & 6.14 & 4.04 & 1.90 & 1.88 & 0.55
& 2.19\% & 5.22\% & 16.53 & 10.04 & 15.36 \\

&
& Dual-Cam (Global)
& 0.52 & 0.75 & 0.76 & 0.27 & \textbf{0.08} & \textbf{0.15}
& \textbf{0.07\%} & 2.22\% & 5.62 & 1.40 & 3.00 \\

&
& US + Dual-Cam
& \textbf{0.25} & \textbf{0.30} & \textbf{0.59} & \textbf{0.17} & 0.10 & 0.25
& 0.11\% & \textbf{1.88\%} & \textbf{3.64} & \textbf{0.79} & \textbf{2.23} \\

& \multirow{3}{*}{S-shaped}
& US only
& 5.56 & 3.46 & 4.27 & 1.96 & 2.46 & 2.33
& 2.20\% & 5.68\% & 16.46 & 8.48 & 16.03 \\

&
& Dual-Cam (Global)
& 0.78 & 0.49 & 0.79 & 0.21 & 0.16 & 0.29
& 0.09\% & \textbf{1.54\%} & 5.40 & 1.43 & 3.85 \\

&
& US + Dual-Cam
& \textbf{0.28} & \textbf{0.42} & \textbf{0.57} & \textbf{0.18} & \textbf{0.11} & \textbf{0.19}
& \textbf{0.01\%} & 1.55\% & \textbf{4.95} & \textbf{0.88} & \textbf{3.79} \\

& \multirow{3}{*}{Zigzag}
& US only
& 2.28 & 4.47 & 10.10 & 2.16 & 2.58 & 1.20
& 2.20\% & 4.10\% & 23.69 & 11.66 & 19.27 \\

&
& Dual-Cam (Global)
& 0.97 & 0.41 & 0.78 & 0.28 & \textbf{0.10} & 0.39
& \textbf{0.06\%} & \textbf{2.06\%} & 5.75 & 1.48 & 5.32 \\

&
& US + Dual-Cam
& \textbf{0.60} & \textbf{0.24} & \textbf{0.47} & \textbf{0.21} & 0.16 & \textbf{0.15}
& \textbf{0.06\%} & 2.07\% & \textbf{5.00} & \textbf{0.88} & \textbf{5.00} \\

\midrule

\multirow{12}{*}{FUSION-L}
& \multirow{3}{*}{C-shaped}
& US only
& 11.79 & 1.48 & 3.97 & 2.71 & 11.53 & 3.21
& 0.96\% & 8.78\% & 26.42 & 12.85 & 24.48 \\

&
& Dual-Cam (Global)
& 0.71 & 0.87 & 0.47 & 0.28 & \textbf{0.09} & 0.22
& \textbf{0.12\%} & 1.66\% & 2.71 & 1.34 & \textbf{2.24} \\

&
& US + Dual-Cam
& \textbf{0.34} & \textbf{0.28} & \textbf{0.33} & \textbf{0.08} & 0.12 & \textbf{0.13}
& 0.24\% & \textbf{0.97\%} & \textbf{1.76} & \textbf{0.64} & 4.43 \\

& \multirow{3}{*}{Linear}
& US only
& 2.79 & 1.23 & 3.60 & 1.19 & 2.58 & 1.57
& 1.08\% & 5.47\% & 13.10 & 5.10 & 6.89 \\

&
& Dual-Cam (Global)
& 0.48 & 0.73 & 1.13 & 0.15 & 0.09 & \textbf{0.09}
& \textbf{0.05\%} & 3.00\% & 4.65 & 1.59 & 4.04 \\

&
& US + Dual-Cam
& \textbf{0.34} & \textbf{0.21} & \textbf{0.43} & \textbf{0.06} & \textbf{0.08} & \textbf{0.09}
& \textbf{0.05\%} & \textbf{2.75\%} & \textbf{3.07} & \textbf{0.68} & \textbf{2.77} \\

& \multirow{3}{*}{S-shaped}
& US only
& 4.93 & 3.50 & 2.81 & 4.35 & 7.59 & 8.69
& 2.55\% & 5.24\% & 13.56 & 7.48 & 11.62 \\

&
& Dual-Cam (Global)
& 0.84 & 0.77 & 1.02 & \textbf{0.15} & \textbf{0.14} & 0.23
& 0.14\% & 2.45\% & 4.79 & 1.72 & 3.24 \\

&
& US + Dual-Cam
& \textbf{0.30} & \textbf{0.36} & \textbf{0.43} & 0.16 & 0.21 & \textbf{0.20}
& \textbf{0.11\%} & \textbf{1.10\%} & \textbf{1.76} & \textbf{0.73} & \textbf{1.76} \\

& \multirow{3}{*}{Zigzag}
& US only
& 7.83 & 1.60 & 7.04 & 2.99 & 14.06 & 9.12
& 3.41\% & 3.27\% & 31.31 & 12.01 & 24.41 \\

&
& Dual-Cam (Global)
& {1.07} & {0.45} & {0.51} & {0.10} & \textbf{0.09} & \textbf{0.15}
& {0.12\%} & {0.62\%} & {2.68} & {1.37} & {2.68} \\

&
& US + Dual-Cam
& \textbf{0.82} & \textbf{0.21} & \textbf{0.34} & \textbf{0.07} & {0.12} & {0.24}
& \textbf{0.11\%} & \textbf{0.50\%} & \textbf{2.46} & \textbf{0.98} & \textbf{2.31} \\

\bottomrule
\end{tabular}
}
\end{table*}

The quantitative results are summarized in Table~\ref{tab:single_traj_comp}. Across the four trajectory shapes, the proposed US + Dual-Cam model consistently outperformed the US-only model by a large margin on both FUSION-J and FUSION-L. Compared with Dual-Cam, the fusion model also consistently reduced AD and MD across all sweep patterns while improving most pose components. For example, AD decreased from 3.11 to 1.75~mm for the C-shaped sweep on FUSION-J and from 1.72 to 0.73~mm for the S-shaped sweep on FUSION-L. Although Dual-Cam remained competitive on a few individual metrics, particularly FDR and ADR, the overall results show that the proposed US + Dual-Cam model provides more stable and accurate pose estimation across representative freehand scanning trajectories.


\subsection{Freehand 3-D US Reconstruction Analysis}
\label{subsubsec:reconstruction_results}

\subsubsection{Reconstruction Method}

Using the full-sweep probe trajectories recovered above, we reconstructed the 3-D US volumes with the voxel-based spatial compounding method of \citet{rohling1997three}. For each scan, every 2-D B-mode frame was transformed into a common world coordinate system using its predicted probe pose. The intensity of each transformed pixel was distributed to neighboring voxels by trilinear interpolation, and the contributions from all frames were accumulated. The final intensity at each voxel was then obtained by normalizing the accumulated intensity by its total interpolation weight. The corresponding 2-D artery segmentation masks were transformed using the same frame-wise poses and compounded on the same voxel grid. This process produced a 3-D label volume spatially aligned with the reconstructed B-mode volume, from which the arterial surface was extracted using a fixed isovalue threshold.

\subsubsection{3-D Reconstruction Results}

We evaluated reconstruction performance on both \textit{in vivo} forearm arteries acquired under the four sweep patterns and phantom vessels acquired along linear trajectories. For each scan, the reconstruction generated using the predicted probe trajectory was compared with a reference generated from the same artery segmentations using ground-truth probe poses. Reconstruction quality was assessed using surface Hausdorff distance ($\mathrm{HD}_{\mathrm{surf}}$) and Chamfer distance ($d_C$) on the extracted surfaces, together with Dice coefficient ($D$) and Jaccard index ($J$) on the thresholded binary volumes. Lower $\mathrm{HD}_{\mathrm{surf}}$ and $d_C$, together with higher $D$ and $J$, indicate better reconstruction fidelity. Formal definitions are provided in Appendix \ref{supp:reconstruction_metrics}.

\begin{figure*}[htp!]
\centering
\includegraphics[width=0.85\textwidth]{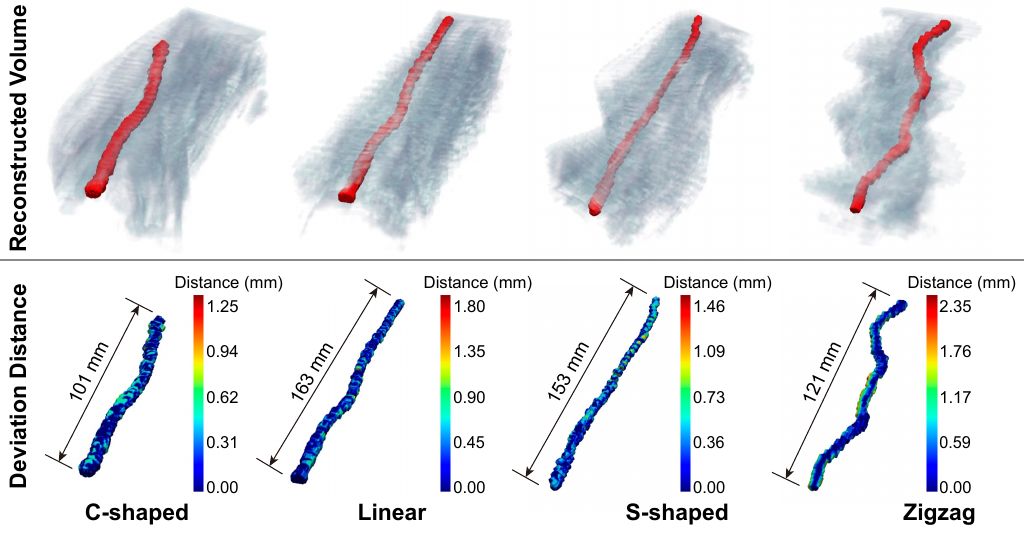}
\caption{\textit{In vivo} 3-D reconstruction results of a forearm artery under four scanning trajectory patterns. The top row shows the reconstructed artery in red and the compounded ultrasound volume in blue-gray. The bottom row shows the point-wise surface distance errors relative to the reference reconstruction obtained using ground-truth probe poses.
}
\label{fig:vessel}
\end{figure*}

\begin{figure}[htp!]
\centering
\includegraphics[width=0.48\textwidth]{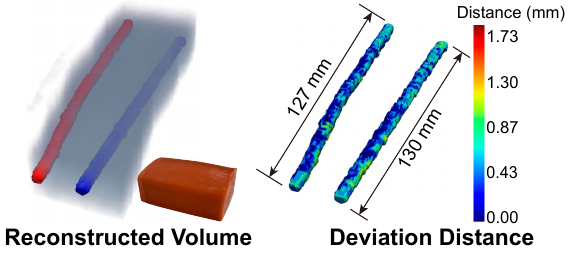}
\caption{Representative 3-D reconstruction results for phantom scans acquired along a linear trajectory.}
\label{fig:phantom}
\end{figure}

Representative qualitative results for the \textit{in vivo} and phantom scans are shown in Figs.~\ref{fig:vessel} and~\ref{fig:phantom}, respectively. The \textit{in vivo} reconstructions preserve the overall arterial geometry across the four sweep patterns, while the phantom reconstructions closely reproduce the reference phantom shapes. The surface-distance maps for the \textit{in vivo} scans further show that the remaining errors are predominantly localized rather than associated with large global distortions. The quantitative results in Table~\ref{tab:reconstruction_metrics} support these observations: across all \textit{in vivo} and phantom reconstructions, $\mathrm{HD}_{\mathrm{surf}}$ ranged from 1.58 to 2.35~mm and $D$ from 0.69 to 0.87. Among the \textit{in vivo} scans, the C-shaped, linear, and S-shaped trajectories achieved stronger geometric agreement than the more challenging zigzag trajectory, whereas the phantom reconstructions remained within the same overall performance range. These findings indicate that the estimated trajectories support geometrically consistent 3-D reconstruction across different sweep patterns and imaging targets.

\begin{table*}[htp!]
\centering
\caption{Quantitative reconstruction results for \textit{in vivo} forearm artery scans and phantom scans.}
\label{tab:reconstruction_metrics}
\small
\setlength{\tabcolsep}{5pt}
\renewcommand{\arraystretch}{1}
\begin{tabular}{cccccc}
\toprule
Scanned Object & Trajectory & $\mathrm{HD}_{\mathrm{surf}}$ (mm) $\downarrow$ & $d_C$ (mm) $\downarrow$ & $D$ $\uparrow$ & $J$ $\uparrow$\\
\midrule
\multirow{4}{*}{\shortstack[c]{\textit{In Vivo}\\Forearm Artery}}
& C-shaped & 1.58 & 0.24 & 0.87 & 0.77\\
& Linear & 1.80 & 0.28 & 0.85 & 0.73\\
& S-shaped & 1.75 & 0.33 & 0.80 & 0.67\\
& Zigzag & 2.35 & 0.58 & 0.69 & 0.53\\
\midrule
Phantom L & \multirow{2}{*}{Linear} & 1.73 & 0.42 & 0.83 & 0.71\\
Phantom R &  & 1.84 & 0.45 & 0.82 & 0.69\\
\bottomrule
\end{tabular}
\end{table*}

\section{Discussion}
\label{sec:discussion}
This study set out to answer not merely whether external visual observations and ultrasound image-domain cues can be combined, but why their combination is principled rather than opportunistic. Two findings frame the discussion. First, ultrasound-only pose estimation benefits from stronger motion-aware feature extraction, yet its long-range accuracy remains constrained by the limited observability of out-of-plane motion in B-mode images. Second, a low-cost dual-camera setup supplies a globally stable pose reference, and treating ultrasound as a source of residual correction, rather than as the primary estimator, is what unlocks the accuracy of both modalities together.

The ultrasound-only results in Table~\ref{tab:dataset1} show that the proposed ultrasound branch matches or exceeds existing sensorless baselines, particularly on trajectory-level metrics. Yet the drift relative to camera-based methods in Table~\ref{tab:ablation_modalities} remains substantially larger, and this gap is revealing: the bottleneck is not network capacity but the information physically available in B-mode images. Out-of-plane motion is encoded only implicitly, through speckle decorrelation that is readily confounded by acoustic shadowing, tissue deformation, and varying probe contact. Ultrasound is therefore locally informative but globally ambiguous: it can sense fine probe-tissue interaction, but it cannot anchor a long sweep in space. Recent advanced temporal models can improve local motion cues, yet they cannot fully recover a global spatial reference that is not directly observable from B-mode images alone.

This directly motivates an external pose reference. In Table~\ref{tab:ablation_modalities}, both single- and dual-camera models reduce pose error and trajectory drift by roughly an order of magnitude relative to the ultrasound-only model, and the dual-camera configuration improves further over either single view. Multi-view geometry resolves the view-dependent ambiguity that a single camera cannot, yielding a markedly more robust estimate of probe motion. Compared with conventional electromagnetic or optical trackers, this dual-camera setup is low-cost, compact, and can be integrated into the freehand scanning setup with relatively low hardware complexity, offering strong pose stability without a proportional rise in system complexity. 

Crucially, our framework does not treat this stability as a reason to discard ultrasound. Instead of replacing one modality with the other, it assigns each the role dictated by its physics: the camera branch provides an externally anchored global trajectory, and the ultrasound branch supplies the local residual that vision is inherently blind to. The consistent gains of US~+~Dual-Cam over Dual-Cam alone across FDR, ADR, MD, AD, and HD on both FUSION-J and FUSION-L confirm that B-mode images retain genuine local motion information, tied to tissue structure, speckle variation, and probe-tissue interaction, that are not directly accessible from external camera observations.

These results point to a dual-modal direction whose value lies in complementarity rather than redundancy. Purely ultrasound-based methods are appealing for their hardware-free simplicity, but their accuracy is capped by the intrinsic ambiguity of B-mode images; purely camera-based methods are trajectory-stable but blind to internal anatomical and speckle-based cues. The proposed residual fusion framework integrates these complementary strengths within a single global-to-local design: vision anchors the trajectory, while ultrasound refines it locally, leading to more accurate probe pose estimation and more geometrically consistent 3-D reconstruction.

Several limitations remain. As full-sequence ultrasound processing is memory-intensive, long trajectories are currently recovered by stitching short overlapping clips rather than modeling the entire sweep end-to-end. A natural next step is a memory-efficient global-local design that distills compact ultrasound motion tokens from local clips and fuses them with full-sweep camera pose sequences. The visual side likewise depends on calibration and visible scene references, and its robustness under marker occlusion and cluttered clinical environments remains to be established. These limitations do not undermine the global-to-local formulation, but indicate important directions for improving system robustness and clinical usability.

\section{Conclusion}
\label{sec:conclusion}

This paper presented a camera-guided residual pose refinement framework that rethinks trackerless freehand 3-D US reconstruction as a global-to-local problem: let vision anchor the trajectory, and let ultrasound refine it. Rather than forcing B-mode images to carry the full burden of pose estimation, the proposed method uses synchronized dual-camera observations to establish an externally anchored, globally stable trajectory, and exploits motion-aware cues from the B-mode sequence to predict only the local residual corrections.

Extensive experiments confirm that this division of labor is highly effective. Ultrasound-only estimation, even with motion-aware feature extraction, remains fundamentally limited by the ambiguity of B-mode speckle, with trajectory drift exceeding $13$~mm. The dual-camera branch alone offers a far more stable reference, and the proposed US~+~Dual-Cam fusion pushes accuracy further still, reducing average trajectory drift to $1.67$~mm and $1.29$~mm on FUSION-J and FUSION-L, which correspond to $16.50\%$ and $27.12\%$ improvements over a strong dual-camera baseline. These gains translate directly into reconstruction quality: across \textit{in vivo} forearm arteries and phantom vessels, the method attains Hausdorff distances of $1.58$ to $2.35$~mm and Dice scores of $0.69$ to $0.87$ under diverse scanning trajectories.

These results demonstrate that external visual guidance and local ultrasound refinement can result in robust and accurate pose estimation in long-distance or complicated scanning modes. This complementary multi-sensor fusion pipeline offers a low-cost and readily deployable route toward accurate, robust, and more automated freehand 3-D ultrasound reconstruction. We believe it opens a practical path for bringing volumetric US imaging beyond specialized settings into routine clinical and point-of-care use.

\printcredits

\section*{Declaration of competing interest}
\label{sec:competing_interest}
The authors declare that they have no known competing financial interests or personal relationships that could have appeared to influence the work reported in this paper.

\section*{Declaration of generative AI use}
\label{sec:ai_declaration}
During the preparation of this manuscript, the authors used ChatGPT to improve its language and readability. All AI-assisted content was carefully reviewed and edited by the authors. The authors take full responsibility for the final content of the article.

\section*{Data availability}
\label{sec:data_availability}
All code and datasets used in this study will be made publicly available upon acceptance of the manuscript.

\section*{Acknowledgements}
\label{sec:acknowledgements}
This work was supported in part by the Multi-Scale Medical Robotics Center, AIR@InnoHK, Hong Kong. The authors also thank Prof. Wei-Ning Lee for his assistance with data acquisition.

\appendices
\renewcommand{\thesection}{\Alph{section}}

\renewcommand{\theequation}{\thesection.\arabic{equation}}
\setcounter{equation}{0}

\section{Multi-Scale Weighting Strategy}
\label{supp:multiscale_weighting}

Sequential composition causes small frame-wise pose errors to accumulate over time. To motivate the horizon-dependent weight in the multi-scale pose loss, we express each predicted relative transformation as a right-multiplicative perturbation of its ground truth:
\begin{equation}
\Delta\hat{\mathbf{T}}_i
=
\Delta\mathbf{T}_i
\exp(\boldsymbol{\zeta}_i^{\wedge}),
\end{equation}
where $\boldsymbol{\zeta}_i\in\mathbb{R}^{6}$ is the local twist error and $(\cdot)^{\wedge}$ maps it to $\mathfrak{se}(3)$. We decompose this error into a systematic bias and a zero-mean stochastic component:
\begin{equation}
\boldsymbol{\zeta}_i
=
\boldsymbol{\mu}
+
\boldsymbol{\epsilon}_i,
\quad
\mathbb{E}[\boldsymbol{\epsilon}_i]=\mathbf{0},
\quad
\operatorname{Cov}(\boldsymbol{\epsilon}_i)=\Sigma.
\end{equation}
Under the first-order small-error approximation on $SE(3)$, and assuming small inter-frame motions such that the associated tangent-space mappings can be approximated by the identity, the error of a $k$-step composition can be approximated as
\begin{equation}
\boldsymbol{\zeta}_{i}^{(k)}
\approx
\sum_{r=i}^{i+k-1}
\boldsymbol{\zeta}_r
=
k\boldsymbol{\mu}
+
\sum_{r=i}^{i+k-1}\boldsymbol{\epsilon}_r.
\end{equation}

Assuming that the stochastic errors are independent or weakly correlated with finite correlation length, the bias terms add coherently, whereas the stochastic terms accumulate approximately as a random walk. The second moment therefore scales as
\begin{equation}
\mathbb{E}
\left[
\left\|
\boldsymbol{\zeta}_{i}^{(k)}
\right\|_2^2
\right]
=
\mathcal{O}
\left(
k^2\|\boldsymbol{\mu}\|_2^2
+
k\operatorname{tr}(\Sigma)
\right).
\end{equation}
Consequently, systematic bias grows as $\mathcal{O}(k)$ in root-mean-square magnitude, whereas stochastic error grows as $\mathcal{O}(\sqrt{k})$. Equal weighting would therefore allow long-horizon stochastic variation to grow with $k$, while overly aggressive down-weighting would suppress the desired sensitivity to systematic drift. We choose the intermediate normalization
\begin{equation}
w_k=\frac{1}{\sqrt{k}}.
\end{equation}
At the root-mean-square error level, this choice approximately normalizes the stochastic contribution across temporal horizons because $w_k\sqrt{k}=\mathcal{O}(1)$, while retaining greater sensitivity to systematic drift because $w_k k=\mathcal{O}(\sqrt{k})$.

\section{Pose and Reconstruction Evaluation Metrics}
\label{supp:reconstruction_metrics}
\setcounter{equation}{0}

Let $\boldsymbol{\theta}_i$ and $\hat{\boldsymbol{\theta}}_i$ denote the ground-truth and predicted global pose vectors obtained from $\mathbf{T}_i$ and $\hat{\mathbf{T}}_i$, respectively. The MAE of component $d\in\{x,y,z,\mathrm{roll},\mathrm{pitch},\mathrm{yaw}\}$ is
\begin{equation}
\mathrm{MAE}_{d}
=
\frac{1}{N}
\sum_{i=0}^{N-1}
\left|
\hat{\theta}_{i,d}
-
\theta_{i,d}
\right|.
\end{equation}

For trajectory metrics, let $\mathbf{p}_i=\operatorname{trans}(\mathbf{T}_i)$ and $\hat{\mathbf{p}}_i=\operatorname{trans}(\hat{\mathbf{T}}_i)$ be the reference and predicted positions after alignment to their first frames. The point-wise position error and cumulative reference path length are
\begin{equation}
e_i=\|\hat{\mathbf{p}}_i-\mathbf{p}_i\|_2,
\quad
\ell_i=\sum_{j=1}^{i}\|\mathbf{p}_j-\mathbf{p}_{j-1}\|_2.
\end{equation}
The FDR, ADR, MD, and AD are defined as
\begin{equation}
\mathrm{FDR}=\frac{e_{N-1}}{\ell_{N-1}}\times100\%,
\quad
\mathrm{ADR}=\frac{1}{|\mathcal{I}|}\sum_{i\in\mathcal{I}}\frac{e_i}{\ell_i}\times100\%,
\end{equation}
\begin{equation}
\mathrm{MD}=\max_{0\leq i\leq N-1}e_i,
\quad
\mathrm{AD}=\frac{1}{N}\sum_{i=0}^{N-1}e_i,
\end{equation}
where $\mathcal{I}=\{i\mid\ell_i>\tau\}$ excludes near-zero path lengths. The trajectory Hausdorff distance measures the worst-case discrepancy between the sampled trajectory point sets $\mathcal{P}=\{\mathbf{p}_i\}_{i=0}^{N-1}$ and $\hat{\mathcal{P}}=\{\hat{\mathbf{p}}_i\}_{i=0}^{N-1}$:
\begin{equation}
\begin{aligned}
\mathrm{HD}_{\mathrm{traj}}
={}&
\max\left\{
\max_{\mathbf{p}\in\mathcal{P}}\min_{\hat{\mathbf{p}}\in\hat{\mathcal{P}}}\|\mathbf{p}-\hat{\mathbf{p}}\|_2, \right.\\
&\left.
\max_{\hat{\mathbf{p}}\in\hat{\mathcal{P}}}\min_{\mathbf{p}\in\mathcal{P}}\|\hat{\mathbf{p}}-\mathbf{p}\|_2
\right\}.
\end{aligned}
\end{equation}

For reconstruction quality, let $S_P$ and $S_R$ denote the predicted and reference surface point sets, respectively, and let $V_P$ and $V_R$ denote the corresponding thresholded binary voxel sets.

The surface Hausdorff distance, $\mathrm{HD}_{\mathrm{surf}}$, measures the worst-case geometric discrepancy between the predicted and reference surfaces:
\begin{equation}
\begin{aligned}
\mathrm{HD}_{\mathrm{surf}}(S_P,S_R)
= {} & \max\left\{\max_{s_p\in S_P}\min_{s_r\in S_R}\|s_p-s_r\|, \right.\\
& \left.\max_{s_r\in S_R}\min_{s_p\in S_P}\|s_r-s_p\|\right\}.
\end{aligned}
\end{equation}

The Chamfer distance ($d_C$) measures the average bidirectional surface discrepancy:
\begin{equation}
\begin{aligned}
d_C(S_P,S_R)
= {} & \frac{1}{|S_P|}\sum_{s_p\in S_P}\min_{s_r\in S_R}\|s_p-s_r\|\\
& +\frac{1}{|S_R|}\sum_{s_r\in S_R}\min_{s_p\in S_P}\|s_r-s_p\|.
\end{aligned}
\end{equation}

The Dice coefficient ($D$) measures volumetric overlap in the thresholded 3-D binary voxel space:
\begin{equation}
D(V_P,V_R)=\frac{2|V_P\cap V_R|}{|V_P|+|V_R|}.
\end{equation}

The Jaccard index ($J$) measures the intersection-over-union ratio in the same space:
\begin{equation}
J(V_P,V_R)=\frac{|V_P\cap V_R|}{|V_P\cup V_R|}.
\end{equation}

Lower $\mathrm{HD}_{\mathrm{traj}}$, $\mathrm{HD}_{\mathrm{surf}}$, and $d_C$, together with higher $D$ and $J$, indicate better trajectory or reconstruction fidelity.

\bibliographystyle{cas-model2-names}
\bibliography{references}

@article{men2023gaze,
  title={Gaze-probe joint guidance with multi-task learning in obstetric ultrasound scanning},
  author={Men, Qianhui and Teng, Clare and Drukker, Lior and Papageorghiou, Aris T and Noble, J Alison},
  journal={Medical image analysis},
  volume={90},
  pages={102981},
  year={2023},
  publisher={Elsevier}
}

@article{tehrani2022lateral,
  title={Lateral strain imaging using self-supervised and physically inspired constraints in unsupervised regularized elastography},
  author={Tehrani, Ali KZ and Ashikuzzaman, Md and Rivaz, Hassan},
  journal={IEEE Transactions on Medical Imaging},
  volume={42},
  number={5},
  pages={1462--1471},
  year={2022},
  publisher={IEEE}
}

@article{rohling1997three,
  title={Three-dimensional spatial compounding of ultrasound images},
  author={Rohling, Robert and Gee, Andrew and Berman, Laurence},
  journal={Medical Image Analysis},
  volume={1},
  number={3},
  pages={177--193},
  year={1997},
  publisher={Elsevier}
}

@inproceedings{ravi2025sam,
  title={Sam 2: Segment anything in images and videos},
  author={Ravi, Nikhila and Gabeur, Valentin and Hu, Yuan-Ting and Hu, Ronghang and Ryali, Chaitanya and Ma, Tengyu and Khedr, Haitham and R{\"a}dle, Roman and Rolland, Chloe and Gustafson, Laura and others},
  booktitle={International Conference on Learning Representations},
  volume={2025},
  pages={28085--28128},
  year={2025}
}

@inproceedings{wen2024foundationpose,
  title={Foundationpose: Unified 6d pose estimation and tracking of novel objects},
  author={Wen, Bowen and Yang, Wei and Kautz, Jan and Birchfield, Stan},
  booktitle={Proceedings of the IEEE/CVF conference on computer vision and pattern recognition},
  pages={17868--17879},
  year={2024}
}

@article{zhang2026mlrecon,
  title={MLRecon: Robust Markerless Freehand 3D Ultrasound Reconstruction via Coarse-to-Fine Pose Estimation},
  author={Zhang, Yi and Tu, Puxun and Wang, Kun and Yan, Yulin and Ying, Tao and Chen, Xiaojun},
  journal={arXiv preprint arXiv:2603.00990},
  year={2026}
}

@article{liang2026sensorless,
  title={Sensorless 3D Ultrasound Reconstruction via Embedded Patterns: A Low-Cost Framework},
  author={Liang, Libin and Dai, Ling and Xing, Shuwei and Zhao, Kaitao and Li, Zhongyu and Zhu, Jihua and Fenster, Aaron},
  journal={IEEE Transactions on Instrumentation and Measurement},
  year={2026},
  publisher={IEEE}
}

@article{hu2023wearable,
  title={A wearable cardiac ultrasound imager},
  author={Hu, Hongjie and Huang, Hao and Li, Mohan and Gao, Xiaoxiang and Yin, Lu and Qi, Ruixiang and Wu, Ray S and Chen, Xiangjun and Ma, Yuxiang and Shi, Keren and others},
  journal={Nature},
  volume={613},
  number={7945},
  pages={667--675},
  year={2023},
  publisher={Nature Publishing Group UK London}
}

@article{li2025tus,
  title={TUS-REC2024: A challenge to reconstruct 3D freehand ultrasound without external tracker},
  author={Li, Qi and Saeed, Shaheer U and Huang, Yuliang and Luo, Mingyuan and Yan, Zhongnuo and Chen, Jiongquan and Yang, Xin and Ni, Dong and Winter, Nektarios and Nguyen, Phuc and others},
  journal={arXiv preprint arXiv:2506.21765},
  year={2025}
}

@article{jiang2023dopus,
  title={Dopus-net: Quality-aware robotic ultrasound imaging based on doppler signal},
  author={Jiang, Zhongliang and Duelmer, Felix and Navab, Nassir},
  journal={IEEE Transactions on Automation Science and Engineering},
  volume={21},
  number={3},
  pages={3229--3242},
  year={2023},
  publisher={IEEE}
}

@article{huang2024robot,
  title={Robot-assisted deep venous thrombosis ultrasound examination using virtual fixture},
  author={Huang, Dianye and Yang, Chenguang and Zhou, Mingchuan and Karlas, Angelos and Navab, Nassir and Jiang, Zhongliang},
  journal={IEEE Transactions on Automation Science and Engineering},
  volume={22},
  pages={381--392},
  year={2024},
  publisher={IEEE}
}

@article{wang2023virtual,
  title={A virtual ultrasonography simulator for skill training using magnetic-inertial probe tracking},
  author={Wang, Heng and Dong, Shuda and Yang, Qi and Han, Jiancheng and He, Ze'an and He, Yihua and Wang, Shuangyi},
  journal={IEEE/ASME Transactions on Mechatronics},
  volume={29},
  number={1},
  pages={445--454},
  year={2023},
  publisher={IEEE}
}

@article{li2023automatic,
  title={Automatic diagnosis of carotid atherosclerosis using a portable freehand 3-D ultrasound imaging system},
  author={Li, Jiawen and Huang, Yunqian and Song, Sheng and Chen, Hongbo and Shi, Junni and Xu, Duo and Zhang, Haibin and Chen, Man and Zheng, Rui},
  journal={IEEE transactions on ultrasonics, ferroelectrics, and frequency control},
  volume={71},
  number={2},
  pages={266--279},
  year={2023},
  publisher={IEEE}
}

@article{verhoef2025freehand,
  title={Freehand ultrafast Doppler ultrasound imaging with optical tracking allows for detailed 3D reconstruction of blood flow in the human brain},
  author={Verhoef, Luuk and Soloukey, Sadaf and Mastik, Frits and Generowicz, Bastian S and Bos, Eelke M and Schouten, Joost W and Koekkoek, Sebastiaan KE and Vincent, Arnaud JPE and Klein, Stefan and Kruizinga, Pieter},
  journal={IEEE Transactions on Medical Imaging},
  year={2025},
  publisher={IEEE}
}

@article{wang2023multiplexed,
  title={A multiplexed 32$\times$ 32 2d matrix array transducer for flexible sub-aperture volumetric ultrasound imaging},
  author={Wang, Ningyuan and Qiang, Yu and Qiu, Chaorui and Chen, Yiliang and Wang, Xingying and Pan, Yue and Liu, Rong and Wu, Weichang and Zheng, Hairong and Qiu, Weibao and others},
  journal={IEEE Transactions on Biomedical Engineering},
  volume={71},
  number={3},
  pages={831--840},
  year={2023},
  publisher={IEEE}
}

@article{adriaans2024trackerless,
  title={Trackerless 3D freehand ultrasound reconstruction: a review},
  author={Adriaans, Chrissy A and Wijkhuizen, Mark and van Karnenbeek, Lennard M and Geldof, Freija and Dashtbozorg, Behdad},
  journal={Applied Sciences},
  volume={14},
  number={17},
  pages={7991},
  year={2024},
  publisher={MDPI}
}

@article{zhang2026navigation,
  title={A navigation-guided 3D breast ultrasound scanning and reconstruction system for automated multi-lesion spatial localization and diagnosis},
  author={Zhang, Yi and Yan, Yulin and Wang, Kun and Cai, Muyu and Xiang, Yifei and Guo, Yan and Tu, Puxun and Ying, Tao and Chen, Xiaojun},
  journal={Medical Image Analysis},
  pages={103965},
  year={2026},
  publisher={Elsevier}
}

@article{bureau2023three,
  title={Three-dimensional ultrasound matrix imaging},
  author={Bureau, Flavien and Robin, Justine and Le Ber, Arthur and Lambert, William and Fink, Mathias and Aubry, Alexandre},
  journal={Nature Communications},
  volume={14},
  number={1},
  pages={6793},
  year={2023},
  publisher={Nature Publishing Group UK London}
}

@article{huang2025vibnet,
  title={Vibnet: Vibration-boosted needle detection in ultrasound images},
  author={Huang, Dianye and Li, Chenyang and Karlas, Angelos and Chu, Xiangyu and Au, KW Samuel and Navab, Nassir and Jiang, Zhongliang},
  journal={IEEE Transactions on Medical Imaging},
  year={2025},
  publisher={IEEE}
}

@article{yates2025improving,
  title={Improving polymyalgia rheumatica care: considerations for routine vascular ultrasound in clinical practice},
  author={Yates, Max and Davies, Charlotte and MacGregor, Alexander James},
  journal={The Lancet Rheumatology},
  volume={7},
  number={6},
  pages={e448--e450},
  year={2025},
  publisher={Elsevier}
}

@article{komatsu2025establishment,
  title={Establishment of High-Precision ultrasound diagnosis methods based on the introduction of deep learning},
  author={Komatsu, Masaaki and Komatsu, Reina and Sakai, Akira and Yasutomi, Suguru and Harada, Naoaki and Aoyama, Rina and Teraya, Naoki and Takeda, Katsuji and Natsume, Takashi and Taniguchi, Tomonori and others},
  journal={IEEE Reviews in Biomedical Engineering},
  year={2025},
  publisher={IEEE}
}

@inproceedings{wilson2025dualtrack,
  title={Dualtrack: Sensorless 3d ultrasound needs local and global context},
  author={Wilson, Paul FR and Ronchetti, Matteo and G{\"o}bl, R{\"u}diger and Markova, Viktoria and Rosenzweig, Sebastian and Prevost, Raphael and Mousavi, Parvin and Zettinig, Oliver},
  booktitle={International Workshop on Advances in Simplifying Medical Ultrasound},
  pages={3--12},
  year={2025},
  organization={Springer}
}

@inproceedings{ning2022spatial,
  title={Spatial position estimation method for 3d ultrasound reconstruction based on hybrid transfomers},
  author={Ning, Guochen and Liang, Hanying and Zhou, Lei and Zhang, Xinran and Liao, Hongen},
  booktitle={2022 IEEE 19th International Symposium on Biomedical Imaging (ISBI)},
  pages={1--5},
  year={2022},
  organization={IEEE}
}

@article{luo2025monetv2,
  title={MoNetV2: Enhanced Motion Network for Freehand 3-D Ultrasound Reconstruction},
  author={Luo, Mingyuan and Yang, Xin and Yan, Zhongnuo and Cao, Yan and Zhang, Yuanji and Hu, Xindi and Wang, Jin and Ding, Haoxuan and Han, Wei and Sun, Litao and others},
  journal={IEEE Transactions on Neural Networks and Learning Systems},
  year={2025},
  publisher={IEEE}
}

@inproceedings{kendall2018multi,
  title={Multi-task learning using uncertainty to weigh losses for scene geometry and semantics},
  author={Kendall, Alex and Gal, Yarin and Cipolla, Roberto},
  booktitle={Proceedings of the IEEE conference on computer vision and pattern recognition},
  pages={7482--7491},
  year={2018}
}

@inproceedings{huang2025improving,
  title={Improving Probe Localization for Freehand 3D Ultrasound Using Lightweight Cameras},
  author={Huang, Dianye and Navab, Nassir and Jiang, Zhongliang},
  booktitle={2025 IEEE International Conference on Robotics and Automation (ICRA)},
  pages={15472--15478},
  year={2025},
  organization={IEEE}
}

@article{jiang2021autonomous,
  title={Autonomous robotic screening of tubular structures based only on real-time ultrasound imaging feedback},
  author={Jiang, Zhongliang and Li, Zhenyu and Grimm, Matthias and Zhou, Mingchuan and Esposito, Marco and Wein, Wolfgang and Stechele, Walter and Wendler, Thomas and Navab, Nassir},
  journal={IEEE Transactions on Industrial Electronics},
  volume={69},
  number={7},
  pages={7064--7075},
  year={2021},
  publisher={IEEE}
}

@article{jiang2023robotic,
  title={Robotic ultrasound imaging: State-of-the-art and future perspectives},
  author={Jiang, Zhongliang and Salcudean, Septimiu E and Navab, Nassir},
  journal={Medical image analysis},
  pages={102878},
  year={2023},
  publisher={Elsevier}
}

@article{lee2023panoramic,
  title={Panoramic volumetric clinical handheld photoacoustic and ultrasound imaging},
  author={Lee, Changyeop and Cho, Seonghee and Lee, Donghyun and Lee, Jonghun and Park, Jong-Il and Kim, Hong-Ju and Park, Sae Hyun and Choi, Wonseok and Kim, Ung and Kim, Chulhong},
  journal={Photoacoustics},
  volume={31},
  pages={100512},
  year={2023},
  publisher={Elsevier}
}

@article{trahey1986speckle,
  title={Speckle pattern correlation with lateral aperture translation: Experimental results and implications for spatial compounding},
  author={Trahey, Gregg E and Smith, Stephen W and Von Ramm, Olaf T},
  journal={IEEE transactions on ultrasonics, ferroelectrics, and frequency control},
  volume={33},
  number={3},
  pages={257--264},
  year={1986},
  publisher={IEEE}
}

@article{tuthill1998automated,
  title={Automated three-dimensional US frame positioning computed from elevational speckle decorrelation.},
  author={Tuthill, Theresa A and Kr{\"u}cker, JF and Fowlkes, J Brian and Carson, Paul L},
  journal={Radiology},
  volume={209},
  number={2},
  pages={575--582},
  year={1998}
}

@article{chang20033,
  title={3-D US frame positioning using speckle decorrelation and image registration},
  author={Chang, Ruey-Feng and Wu, Wen-Jie and Chen, Dar-Ren and Chen, Wei-Ming and Shu, Wei and Lee, Jau-Hong and Jeng, Long-Bin},
  journal={Ultrasound in medicine \& biology},
  volume={29},
  number={6},
  pages={801--812},
  year={2003},
  publisher={Elsevier}
}

@article{gee2006sensorless,
  title={Sensorless freehand 3D ultrasound in real tissue: Speckle decorrelation without fully developed speckle},
  author={Gee, Andrew H and Housden, R James and Hassenpflug, Peter and Treece, Graham M and Prager, Richard W},
  journal={Medical image analysis},
  volume={10},
  number={2},
  pages={137--149},
  year={2006},
  publisher={Elsevier}
}

@article{laporte2011learning,
  title={Learning to estimate out-of-plane motion in ultrasound imagery of real tissue},
  author={Laporte, Catherine and Arbel, Tal},
  journal={Medical image analysis},
  volume={15},
  number={2},
  pages={202--213},
  year={2011},
  publisher={Elsevier}
}

@article{prevost20183d,
  title={3D freehand ultrasound without external tracking using deep learning},
  author={Prevost, Raphael and Salehi, Mehrdad and Jagoda, Simon and Kumar, Navneet and Sprung, Julian and Ladikos, Alexander and Bauer, Robert and Zettinig, Oliver and Wein, Wolfgang},
  journal={Medical image analysis},
  volume={48},
  pages={187--202},
  year={2018},
  publisher={Elsevier}
}

@inproceedings{guo2020sensorless,
  title={Sensorless freehand 3D ultrasound reconstruction via deep contextual learning},
  author={Guo, Hengtao and Xu, Sheng and Wood, Bradford and Yan, Pingkun},
  booktitle={International Conference on Medical Image Computing and Computer-Assisted Intervention},
  pages={463--472},
  year={2020},
  organization={Springer}
}

@article{guo2022ultrasound,
  title={Ultrasound volume reconstruction from freehand scans without tracking},
  author={Guo, Hengtao and Chao, Hanqing and Xu, Sheng and Wood, Bradford J and Wang, Jing and Yan, Pingkun},
  journal={IEEE Transactions on Biomedical Engineering},
  volume={70},
  number={3},
  pages={970--979},
  year={2022},
  publisher={IEEE}
}

@inproceedings{luo2022deep,
  title={Deep motion network for freehand 3D ultrasound reconstruction},
  author={Luo, Mingyuan and Yang, Xin and Wang, Hongzhang and Du, Liwei and Ni, Dong},
  booktitle={International Conference on Medical Image Computing and Computer-Assisted Intervention},
  pages={290--299},
  year={2022},
  organization={Springer}
}

@article{li2023long,
  title={Long-term dependency for 3d reconstruction of freehand ultrasound without external tracker},
  author={Li, Qi and Shen, Ziyi and Li, Qian and Barratt, Dean C and Dowrick, Thomas and Clarkson, Matthew J and Vercauteren, Tom and Hu, Yipeng},
  journal={IEEE Transactions on Biomedical Engineering},
  volume={71},
  number={3},
  pages={1033--1042},
  year={2023},
  publisher={IEEE}
}

@inproceedings{yan2024fine,
  title={Fine-grained context and multi-modal alignment for freehand 3d ultrasound reconstruction},
  author={Yan, Zhongnuo and Yang, Xin and Luo, Mingyuan and Chen, Jiongquan and Chen, Rusi and Liu, Lian and Ni, Dong},
  booktitle={International Conference on Medical Image Computing and Computer-Assisted Intervention},
  pages={340--349},
  year={2024},
  organization={Springer}
}

@article{dou2024sensorless,
  title={Sensorless end-to-end freehand 3-D ultrasound reconstruction with physics-guided deep learning},
  author={Dou, Yimeng and Mu, Fangzhou and Li, Yin and Varghese, Tomy},
  journal={IEEE transactions on ultrasonics, ferroelectrics, and frequency control},
  volume={71},
  number={11},
  pages={1514--1525},
  year={2024},
  publisher={IEEE}
}

@article{lee2025enhancing,
  title={Enhancing free-hand 3d photoacoustic and ultrasound reconstruction using deep learning},
  author={Lee, SiYeoul and Kim, Seonho and Seo, Minkyung and Park, SeongKyu and Imrus, Salehin and Ashok, Kambaluru and Lee, DongEon and Park, Chunsu and Lee, SeonYeong and Kim, Jiye and others},
  journal={IEEE Transactions on Medical Imaging},
  year={2025},
  publisher={IEEE}
}

@inproceedings{sun2014probe,
  title={Probe localization for freehand 3D ultrasound by tracking skin features},
  author={Sun, Shih-Yu and Gilbertson, Matthew and Anthony, Brian W},
  booktitle={International Conference on Medical Image Computing and Computer-Assisted Intervention},
  pages={365--372},
  year={2014},
  organization={Springer}
}

@article{zhang2025freehand,
  title={Freehand 3-D Ultrasound Imaging: Sim-in-the-Loop Probe Pose Optimization via Visual Servoing},
  author={Zhang, Yameng and Huang, Dianye and Meng, Max Q-H and Navab, Nassir and Jiang, Zhongliang},
  journal={IEEE/ASME Transactions on Mechatronics},
  year={2025},
  publisher={IEEE}
}

\end{document}